\documentclass{article}

\usepackage{arxiv}

\usepackage[utf8]{inputenc} 
\usepackage[T1]{fontenc}    
\usepackage{hyperref}       
\usepackage{url}            
\usepackage{booktabs}       
\usepackage{amsfonts}       
\usepackage{nicefrac}       
\usepackage{microtype}      
\usepackage{lipsum}

\usepackage{graphicx}
\usepackage{amssymb}
\usepackage{amsthm}
\usepackage{amsmath}

\usepackage[numbers,sort&compress ]{natbib}
\usepackage{float}
\usepackage[table]{xcolor}
\usepackage{xcolor}
\usepackage{pdflscape}
\usepackage{multirow}
\usepackage{subcaption}
\usepackage{array}
\newcolumntype{C}[1]{>{\centering\arraybackslash}p{#1}}

\title{Segmentation of structural parts of rosebush plants with 3D point-based deep learning methods}

\author{
  Kaya Turgut \\
  Department of Electrical-Electronics Engineering\\
  Eskisehir Osmangazi University\\
  Eskisehir, Turkey \\
  \texttt{kayaturgut@hotmail.com} \\
   \And
  Helin Dutagaci \\
  Department of Electrical-Electronics Engineering\\
  Eskisehir Osmangazi University\\
  Eskisehir, Turkey \\
  \texttt{hdutagaci@ogu.edu.tr} \\
   \And
   Gilles Galopin \\
   INRAe, UMR1345 \\
   Institut de Recherche en Horticulture et Semences \\ Angers, France
   \AND
   David Rousseau \\
   Laboratoire Angevin de Recherche en Ingénierie des Systèmes (LARIS) \\
   Université d'Angers \\
   Angers, France \\
   \texttt{david.rousseau@univ-angers.fr}
}

\begin{document}
\maketitle

\begin{abstract}
Segmentation of structural parts of 3D models of plants is an important step for plant phenotyping, especially for monitoring architectural and morphological traits. Current state-of-the art approaches rely on hand-crafted 3D local features for modeling geometric variations in plant structures. While recent advancements in deep learning on point clouds have the potential of extracting relevant local and global characteristics, the scarcity of labeled 3D plant data impedes the exploration of this potential. We adapted six recent point-based deep learning architectures (PointNet, PointNet++, DGCNN, PointCNN, ShellNet, RIConv) for segmentation of structural parts of rosebush models. We generated 3D synthetic rosebush models to provide adequate amount of labeled data for modification and pre-training of these architectures. To evaluate their performance on real rosebush plants, we used the ROSE-X data set of fully annotated point cloud models. We provided experiments with and without the incorporation of synthetic data to demonstrate the potential of point-based deep learning techniques even with limited labeled data of real plants. The experimental results show that PointNet++ produces the highest segmentation accuracy among the six point-based deep learning methods. The advantage of PointNet++ is that it provides a flexibility in the scales of the hierarchical organization of the point cloud data. Pre-training with synthetic 3D models boosted the performance of all architectures, except for PointNet.  
 
\end{abstract}


\section{Background}
Automatic plant phenotyping based on computer vision techniques has become essential for enabling high throughput experiments in botanical and agricultural research \cite{Minervini2015}. While 2D image-based processing facilitates high-throughput phenotyping, advances in 3D data acquisition and modeling provide precise estimation of traits through full, occlusion-free 3D geometric information of plants \cite{Paulus2014, Gibbs2020}.

Several measurements related to plant phenotyping require segmentation of plant parts, such as branches and individual leaves. Shape-related phenotypical traits of potted ornamental plants are especially important for assessing their visual quality \cite{boumaza2009visual}. Architectural traits can be simple, such as the diameters of branches, the number of internodes and stem length \cite{yan2007qtl}. An extended list of more complex architectural traits for rosebush plants is given in \cite{Li-Marchetti2017}. Examples to such traits are number of axes terminated in a flower bud, number of branching orders, lengths of axes and branching angles.  Estimation of length, width and area of leaves provides information for modeling of rose genotypes \cite{Gao2021rose}. In order to automatically extract these phenotypical traits from acquired 3D plant data, a necessary step is identifying the structural category of each 3D point. After stem, flower and leaf points are identified, further processing can be applied to determine individual organs, such as individual leaves, to extract their statistical and geometric characteristics \cite{Xiang2019morph}. Stem points can be processed to detect branching points, which are fundamental for measuring architectural traits \cite{Ziamtsov2020}.

A large body of research has been conducted in recent decades for organ segmentation of plants using machine learning approaches through 2D images and 3D reconstructions  \cite{Scharr2016,Paproki2012,Elnashef2019,Gelard2017,Wahabzada2015,Li2013,Paulus2013,Hetroywheeler2016,Golbach2016,MVAPPound2016,Liu2020,Mack2020}. The common practice for segmentation of 3D models is to extract hand-crafted local surface features, such as eigenvalues of local covariance matrix \cite{Dey2012} or the second tensor \cite{Elnashef2019}, Fast Point Feature Histograms (FPFH) \cite{Paulus2013,Paulus2014b,Wahabzada2015,Sodhi2017}, and surface curvature \cite{Li2013}. Local features can as well be extracted from volumetric representations of plants. Extraction of eigenvalues of the second-moments tensor of the 3D neighbourhood \cite{Klodt2014}, a breath-first flood-fill algorithm with a 26-connected neighbourhood \cite{Golbach2016}, extraction of multi-scale texture and edge features \cite{RoseX2020} are examples to volumetric approaches. In \cite{Dey2012,Paulus2013,Sodhi2017,RoseX2020}, semantic segmentation methods are equipped with supervised learning techniques such as Support Vector Machines and Random Forests. Markov Random Fields (MRF)-based smoothing over class labels \cite{Li2013,Sodhi2017} or region growing \cite{Paulus2013,Paulus2014b} are occasionally used to ensure consistency of point labels within local regions.

Apart from segmentation methods based on local features, graph-based approaches involving spectral embedding and clustering \cite{Hetroywheeler2016,Santos2015} can also be effective. Another strategy is fitting geometric primitives such as ellipses, tubular structures, cylinders or rings to 3D data for semantic segmentation \cite{Binney2009,Paproki2012,Chaivivatrakul2014,Gelard2017}.

Deep learning methods, in contrast to the use of hand-crafted features, have the advantage of being able to learn features from raw input data and model the within-class and between-class variations of the features simultaneously. Their application to 2D image-based plant detection, phenotyping and part-segmentation have been proven to be successful \cite{Gao2020, Jiang2020,Ubbens2017,Pound2017,MVAPAtanbori2020, MVAPKumar2020, Grimm2019, Samiei2020, Jiang2020PM}. Despite this trend, deep learning methods that directly consume 3D point clouds have not been explored for 3D plant phenotyping. The main factor that impedes this exploration is the requirement for large amount of training data and the lack of large annotated 3D plant data sets \cite{Chaudhury2020}. Even moderate size annotated data sets of full plant models are not available. As opposed to the speed of acquiring and annotating 2D images, the procedures for 3D model reconstruction and annotation of real plants are time-demanding and error-prone. 

A strategy to reduce this time consuming step is using synthetic data generated with their associated ground truth. This approach has been extensively used in plant phenotyping with 2D images
\cite{barth2018data,di2017automatic,frid2018synthetic,valerio2017arigan,pawara2017data,valerio2017arigan,ward2018deep,zhu2018data,douarre2019novel}. Incorporation of synthetic plants through generative models such as Lindenmayer systems (L-systems) \cite{Lindenmayer68,Prusinkiewicz1996} into training data is effective with 2D plant phenotyping \cite{Ubbens2018}. The same scheme of creating synthetic 3D plant models can be applied to supply sufficient training data to machine learning frameworks \cite{Chaudhury2020}. 

Virtual plant modeling has been used in agricultural and plant sciences to simulate plant behaviour and analyze interactions of the plants with their environment \cite{Evers2013,Buck-Sorlin2013,Buck-Sorlin2013b}. Examples to platforms that constructs virtual plant models are L+C modelling language \cite{Karwowski2003, Karwowski2004} and L-Py framework \cite{Boudon2012}, both of which are based on the formalism of L-systems \cite{Lindenmayer68}. Despite the availability of such platforms capable of generating synthetic plants with complex architectures, employing them as 3D training data in the form of point clouds for plant phenotyping is not yet practiced.

Research on deep learning methods that directly consume 3D points clouds as input data exploded since the publication of the pioneering work of Qi et al. \cite{Qi2017}, introducing the PointNet \cite{Guo2020,Liu2019,Griffiths2019}. Guo et al. \cite{Guo2020} provide a recent and comprehensive review on deep learning for point clouds. For semantic part segmentation application only, Guo et al. \cite{Guo2020} compare 30 point-based architectures that have been developed since 2017. It is beyond the scope of this paper to mention all these architectures here. The benchmarks with which these architectures are commonly tested are data sets including indoor scenes (S3DIS \cite{Armeni2016}, ScanNet \cite{Dai2017}) or outdoor urban scenes (Semantic3D \cite{Hackel2017}, Semantic KITTI \cite{Behley2019,Geiger2012}).

Despite the fast progress in research on point-based 3D deep learning techniques, their application on plant sciences and agriculture is limited to very few studies. For example, Wu et al. \cite{Wu2020} modified the PointNet architecture for separating foliage and woody components in terrestrial laser scanning data. In \cite{kang2020}, PointNet was used to estimate the proper grasping pose of apples for autonomous harvesting. In some studies aiming part segmentation of 3D plant models, Convolutional Neural Networks (CNN) were applied to 2D multi-view images and the inferences were back-projected to 3D for post-processing \cite{Shi2019,Japes2018}. In \cite{Jin2020} a  voxel-based convolutional  neural  network  (VCNN)  was designed for maize stem and leaf classification and segmentation. The point clouds were converted to volumetric models before being processed. The authors briefly compared their method to PointNet and PointNet++ in terms of segmentation accuracy. To the best of our knowledge, this is the only work where the authors reported part segmentation results on 3D plant models using point-based deep learning architectures.

Exploration of the performance of recent deep learning techniques on 3D plant phenotyping is imperative since these approaches have the promise of simultaneous extraction of relevant information from the data at various scales and learning to design classifiers that model the variability in the data. They have been proven to outperform classical machine learning methods that rely on hand-crafted features. However, the recently developed 3D point-based deep learning architectures have not previously been analyzed for their suitability for organ segmentation of full 3D plant models.

The objective of this work is to address this lack of analysis and to provide a benchmark for application of 3D point-based deep learning methods to plant part segmentation. The target data set is the recently introduced ROSE-X data set, which includes eleven 3D models of real rosebush models obtained through X-ray imaging \cite{RoseX2020}. The models are fully annotated with three semantic labels: 1) Flower, 2) Leaf, and 3) Stem (branches and petioles). As baseline methods, six recent 3D point-based deep learning architectures were modified with the help of synthetic models and evaluated for the segmentation of real rosebush plants to their structural parts. 

We used a simulator based on L-networks in order to generate 3D synthetic rosebush (Rosa x hybrida) models. Although 3D synthetic plant models were previously utilized for rendering 2D images for 2D deep learning methods, to the best of our knowledge, they were not previously used in full 3D form for directly enriching the 3D training data for deep learning. In addition to providing a first exploration of the potential of various 3D point-based deep networks for plant phenotyping, this work also presents a first investigation of the contribution of 3D synthetic models for modifying and training such networks. This investigation is particularly important for addressing the challenge of limited labeled 3D plant data. 

In summary, the contributions of this work are
\begin{itemize}
    \item a first analysis of the performance of various 3D point-based deep learning techniques on segmentation of structural parts of full 3D models of real plants;
    \item employment of synthetic 3D plant models for adapting and training 3D point-based deep learning networks;
    \item a benchmark for future developments of 3D point-based architectures targeting 3D plant phenotyping.
\end{itemize}

\section{Material and Method} 

We address the application of 3D point-based deep learning segmentation methods to the specific problem of segmentation of 3D plant models to their structural parts. We considered six such architectures for adaptation to the problem and compared their shortcomings and strengths. The architectures are 1) PointNet \cite{Qi2017}, 2) PointNet++ \cite{Qi2017b}, 3) Dynamic Graph CNN (DGCNN) \cite{Wang2019}, 4) PointCNN \cite{Li2018}, 5) ShellNet \cite{Zhang2019}, and 6) RIConv \cite{Zhang2019b}. We employed the recently introduced ROSE-X data set \cite{RoseX2020}, which includes eleven 3D models of real rosebush plants to train and evaluate the networks. The data set is accompanied with ground truth information in the form of point-level labels of the plant shoot corresponding to three classes: 1) Flower, 2) Leaf, and 3) Stem (branches and petioles).

In order to explore the contribution of using synthetic data for modifying and training the networks, we created a data set consisting of 48 synthetic rosebush (Rosa x hybrida) models. The models were generated by a simulator developed by Favre et al. \cite{Favre2007}. The simulator was implemented with L-studio software \cite{Karwowski2004} based on L-systems. The point clouds extracted from the synthetic data are used to modify and pre-train the networks. Using transfer learning \cite{Yosinski2014}, the networks are updated using the training set of point clouds of the ROSE-X data set. The results on the test models from the ROSE-X were compared with those of the default networks trained without the use of the synthetic data.

\subsection{Data sets}

In this study, we utilized two sets of 3D models of rosebush plants. The first set is the ROSE-X data set, which is composed of 11 fully annotated 3D models of real rosebush plants acquired through X-ray scanning. The second is the set of synthetic rosebush models which were generated using the L-studio-based simulator developed by Favre et al. \cite{Favre2007}. The details of the data sets are provided in the following subsections. The ROSE-X data set is open to public use at \cite{DataRoseX}. 

\subsubsection{ROSE-X data set}

The models in the ROSE-X data set were acquired from real rosebush plants using a 3D X-ray imaging system. The volumetric models were fully annotated with manual supervision and then converted to 3D point clouds. The details of the procedure for annotation and the data structure can be found in \cite{RoseX2020}. Each point in a point cloud belongs to one of three organ classes: Leaf, stem, and flower. The petioles between leaflets were also labeled as stem, since they have branch-like structures and their inclusion to the architecture of branches is important for further analysis. 
 
In most 3D phenotyping experiments, especially for plants of complex architecture, the number of annotated 3D models will be limited. Thus, we set the number of real rosebush plants reserved for training as three. The distribution of points to the three classes for these models is given in Table \ref{table:classDist}.
 
\begin{table}[ht]
\centering
\caption{Distribution of classes in 3D rosebush point clouds (\%).}
\footnotesize
\begin{tabular}{{lcccc}}
\hline
 & Leaf & Stem & Flower \\\hline
ROSE-X training models & 84.56 & 10.63 & 4.81 \\\hline
ROSE-X test models &  81.00 & 11.51 & 7.49 \\\hline
Synthetic models for training &  65.80 & 17.95 & 16.25  \\\hline
Synthetic models for validation & 66.84 & 17.14 & 16.02 \\\hline
\end{tabular}
\label{table:classDist} 
\end{table}

Although the data size in terms of the number of real plants is limited, the plants in the data set are of moderately large ones (30 to 50cm in height) and possess complex architectures with significant variations of the shape and organization of organs within a plant. Furthermore, the plant data is partitioned into blocks each of which is separately processed by the deep learning architectures. The point density of the 3D models allows sampling of 4096 points in each block. From the three rosebush plants reserved for training and validation, we extracted 251 blocks, leading to a moderate amount of data for the purposes of training a machine learning algorithm. For the eight real plants reserved for testing, the number of blocks is even higher (525 blocks) allowing a reliable performance assessment of the deep learning architectures.

\subsubsection{Synthetic Rosebush Models}

To create synthetic rosebush (Rosa x hybrida) models, we used a simulation procedure originally developed by Favre et al. \cite{Favre2007}, and updated in \cite{Garbez2015}. The procedure was implemented with the L-studio software \cite{Karwowski2004}, which provides a modular framework for plant development based on the literature on parametric L-systems \cite{Lindenmayer68,  Prusinkiewicz1997}. This framework makes it possible to integrate measurable characteristics associated with individual modules of specific plant species \cite{ Prusinkiewicz1998}. For the synthetic rosebush model of Favre et al. \cite{Favre2007}, such characteristics were derived from observations on real plants. Morphometric measurements (i.e. diameter and length of organs), architectural structures (i.e. leaf formation order) and physiological data were analyzed and integrated into the model. The simulation model of Favre et al. \cite{Favre2007} was further updated in \cite{Garbez2015} with three core architectural parameters: (1) the number of axes; (2) their location or topology; and (3) their morphologic type (short or long), determined from a five-months old crop of pot plants cultivated in a greenhouse under controlled non-restrictive conditions \cite{Morel2009}.

\begin{figure}[ht]
	\centering
	\includegraphics[width=0.6\textwidth]{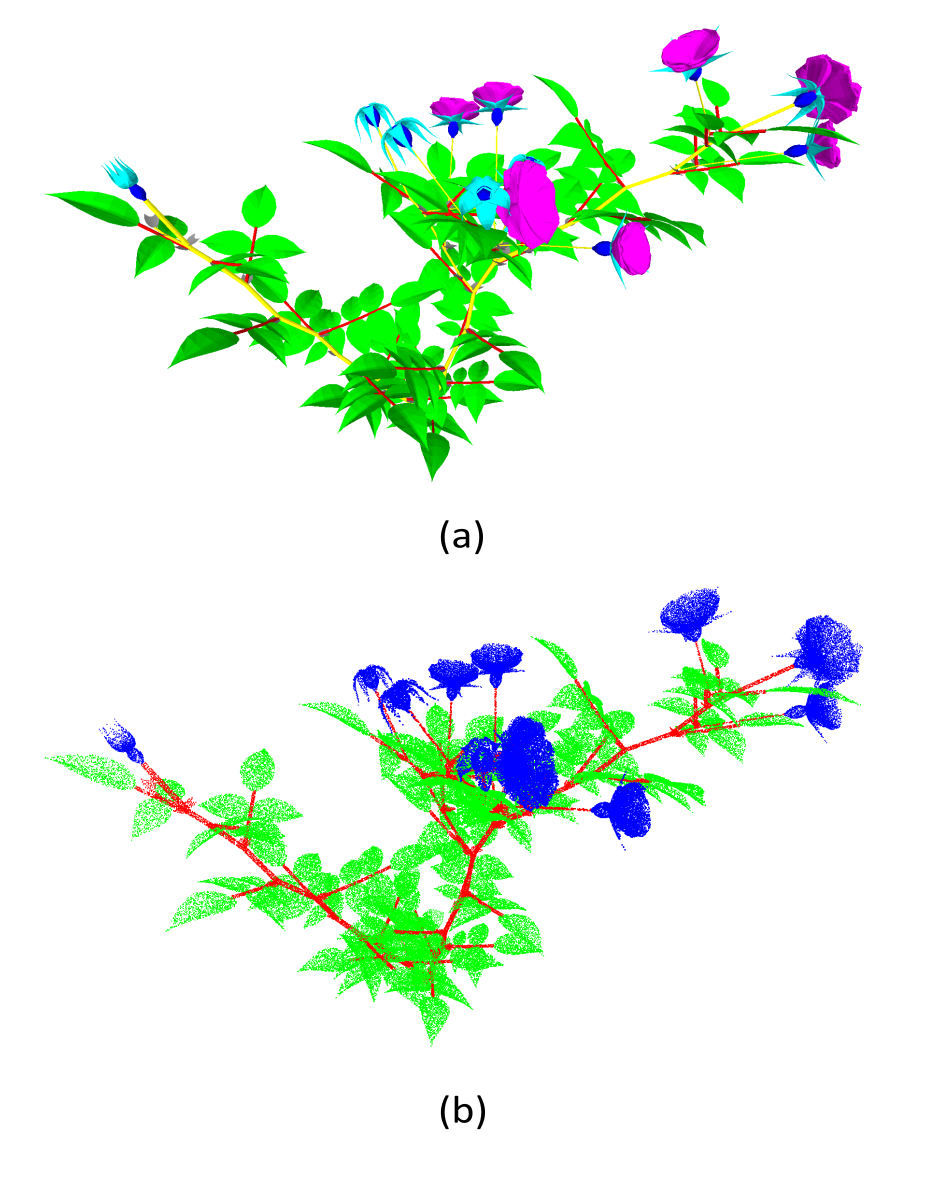}
	\caption{Synthetic plant as a triangle mesh model (a) and the corresponding sampled point cloud (b) }
    \label{fig:virtualroses}
\end{figure}

Using this simulation procedure, we generated 48 different rosebush models in the form of triangle meshes. The triangle mesh and the point cloud of a sample synthetic rosebush model are given in Fig. \ref{fig:virtualroses}. Each triangle in a model is inherently classified into one of seven organs: Leaflet, petiole, stem, stipule, petal, sepal, and receptacle (Fig. \ref{fig:virtualroses}a). Since the ROSE-X labels are not as fine-grained, the petiole, stem and stipule classes were merged together to form the stem class and the sepal, petal and receptacle classes were merged into the flower class after converting the mesh model into a point cloud (Fig. \ref{fig:virtualroses}b).

In order to generate point clouds from these triangle mesh models, we homogeneously sampled points from the triangular surfaces. A point cloud is a set of 3D points $\mathcal{P} = \{p_1,p_2,...,p_N\}$, where each point $p_i \in \mathcal{P}$ is represented with the point's coordinates $(x,y,z)$ in the 3D space. $N$ is the number of points in the $\mathcal{P}$, and it defines the size of the point cloud. The sampling rate was set to 120 points per square unit resulting in point clouds of size of 150,000 to 300,000 points per plant. The dimensions of synthetic models in $x-$, $y-$ and $z-$ axes are in the range of 30 to 50 cm, in accordance to the scale of the real rosebush models. 

For each of the deep learning architectures explored in this paper, we applied many modifications to their default parameters in order to adapt them to segmentation of plants. We modified these parameters experimentally by dividing the synthetic rosebush data into a training and validation set. From the 48 synthetic rosebush models, 8 plants were randomly selected and reserved for validation. The rest of the point clouds are used for training the networks. Similar to the plants in ROSE-X dataset, the synthetic plants are processed through block partitioning. For the total number of blocks extracted from the two sets, please see the Results section.

\subsection{Data preprocessing}
\label{sec:dataprep}

The point-based deep learning architectures accept fixed-size data as input. Feeding the entire rosebush model to the networks requires a large sub-sampling rate resulting in a significant loss of geometric information. Therefore, we follow the strategy commonly used with point-based deep learning methods to handle large-scale point clouds \cite{Li2018}: We partition a rosebush point cloud into fixed-size cubic blocks, each of which is then processed as an independent point cloud by the deep neural networks. The block size in terms of edge length is set as 10 cm through experimentation with the synthetic data set. The networks are trained to segment the organs present in these blocks. At the inference phase, an input plant model is partitioned into blocks, and the predictions from the blocks are combined to obtain a full segmentation. 

In general, the choice of the block size depends on the resolution of the input point cloud. A large cube size will correspond to loss of detail due to subsampling to attain a fixed number of points and a smaller cube will reduce contextual information among semantic parts. Starting from a block size that results in an adequate resolution of the organ surfaces and that covers multiple organs, we varied the block size to increase the performance on the validation set. In our experiments, we found that the performance margin was around 3\% for the networks, by halving or doubling the initial size.

The points in a block should be sampled such that each block includes a fixed number $N$ of points ($N$ is 4096 for the architectures used in this study). We followed a semi-random sampling strategy in order to ensure that the sampled points are distributed in a homogeneous fashion and structures possessing fewer points (like thin branches) are not lost. If there are less than 10\% of $N$ points in a block, the block is discarded and the points in this block are included to a neighboring block. Then, the distribution of the points in each block is analyzed through partitioning the block into voxels with fixed grid size (0.2cm in this work). The average of the number of points in the voxels is calculated. For voxels that have points fewer than the average value, the number of points they contain is increased to the average value by adding copies of the points to the data. Finally, if the points in the block are higher than the allowed number of points, mutually exclusive subsets of $N$ points are selected randomly to form multiple blocks representing the same region. Finally, the blocks with number of points less than $N$ are populated through random point repetition before the training phase. 

To enrich the training data, block partitioning is performed with two different offset values (0 and 5cm) for each training plant model, keeping the block size fixed. In this way, two sets of blocks containing different data from each model are created, providing additional input training data for the networks. 

For segmentation of a new test point cloud, two offset values are used during block partitioning and the blocks of the two sets are fed into the network. As a result, for each point in the point cloud, two sets of probability scores for the part classes are obtained. The class with the highest probability score is assigned to the point. 

\subsection{3D point-based deep learning architectures}

We considered six different 3D point-based deep learning architectures for the problem of part segmentation of rosebush models: 1) PointNet \cite{Qi2017}, 2) PointNet++ \cite{Qi2017b}, 3) Dynamic Graph CNN (DGCNN) \cite{Wang2019}, 4) PointCNN \cite{Li2018}, 5) ShellNet \cite{Zhang2019}, and 6) RIConv \cite{Zhang2019b}. As will be described in detail in Results section, we performed various experiments involving real and synthetic models. We performed extensive experiments with synthetic data alone to modify the architectures in terms of the number of layers, the number of feature channels in the layers, neighborhood sizes, point sampling rates in local neighborhoods, and other hyper-parameters. The final modifications on these parameters correspond to the best-performing settings on the validation set of the synthetic data. The weights of the modified and pre-trained networks are then fine-tuned with real rosebush data. The validation set of real data was instrumental for deciding which weights will be updated during retraining. For the experiments where we excluded synthetic data and used only real models for training, we kept the default settings of the architectures.

In the following subsections, we briefly describe the key approaches of these architectures to the problem of encoding local geometric structure of 3D point clouds. We present the parameters of the architectures that yielded the best performance in the validation set of the synthetic data. For the default structures of the architectures and for other details, please refer to the original articles.

\subsubsection{PointNet}
PointNet architecture \cite{Qi2017} is the first deep neural network architecture that directly accepts a point cloud as input. It uplifts the $(x,y,z)$ coordinates of each 3D point separately to high-dimensional features through Multilayer Perceptrons (MLP) with shared weights. A single maximum pooling operation is applied to summarize all the point features followed by fully-connected (FC) MLPs. The result is a single global feature vector describing the input point cloud. This feature vector is concatenated to individual point-based features to be processed by successive layers. Weight-shared MLP layers are applied to the concatenated features to extract the class scores for each point. 

\begin{figure*}[h]
 \centering
 \includegraphics[width=0.9\textwidth]{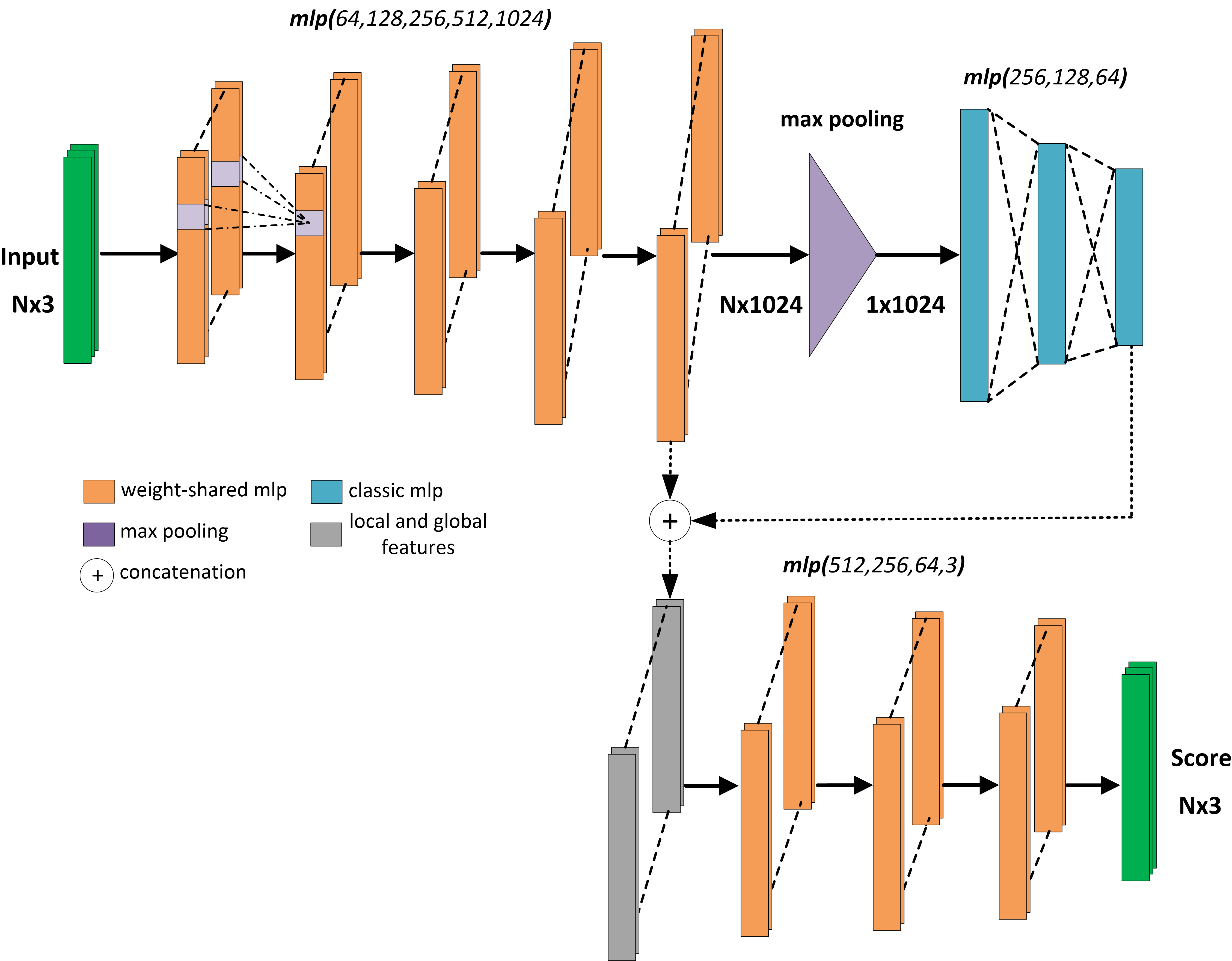}
 \caption{Modified PointNet architecture.}
 \label{fig:PointNetA}
 \end{figure*}

As with other architectures, we modified the default PointNet architecture using the synthetic rosebush models. We inserted an additional FC layer after max-pooling. An additional MLP layer was inserted after the global and point-wise features were concatenated. The number of channels at various layers were also altered. The modified PointNet architecture for segmentation is given in Fig. \ref{fig:PointNetA}. 

PointNet processes each point in an isolated manner upto the max-pooling operation, which generates a global feature vector. The final predictions heavily depend on the locations of the points rather than the local geometric organization around them. There are no connections in the architecture to relate points in close proximity to each other in the Euclidean space. 

\subsubsection{PointNet++}
PointNet++ architecture \cite{Qi2017b} was devised to summarize point-based features in different local scales instead of on the global level. The input point cloud is partitioned into overlapping local regions, and the PointNet is applied to these regions resulting in feature vectors capturing geometric details of local neighbourhoods. Grouping and feature extraction are performed in a hierarchical manner. 

PointNet++ architecture incorporates two types of layers: 1) Set abstraction layer (SA) and 2) Feature propagation layer (FP). SA layer consists of two phases: sampling and grouping. In the sampling phase, $P$ representative points are selected using farthest point sampling algorithm. In the grouping phase, a local neighborhood of fixed radius $R$ is formed around each representative point, resulting in overlapping local groups. In this neighborhood, $M$ points are randomly selected to form a group. PointNet is applied individually to each group to extract features summarized over all the points in the group. FP layers are responsible to propagate the group-based feature vectors to the original points in the input point cloud. The propagation of features to a point is performed via interpolation from the features of its closest neighbours. By combining the interpolated and existing features of SA phase, PointNet architecture is used to update the features of each point. 

 \begin{figure*}[h]
 \centering
 \includegraphics[width=1.0\textwidth]{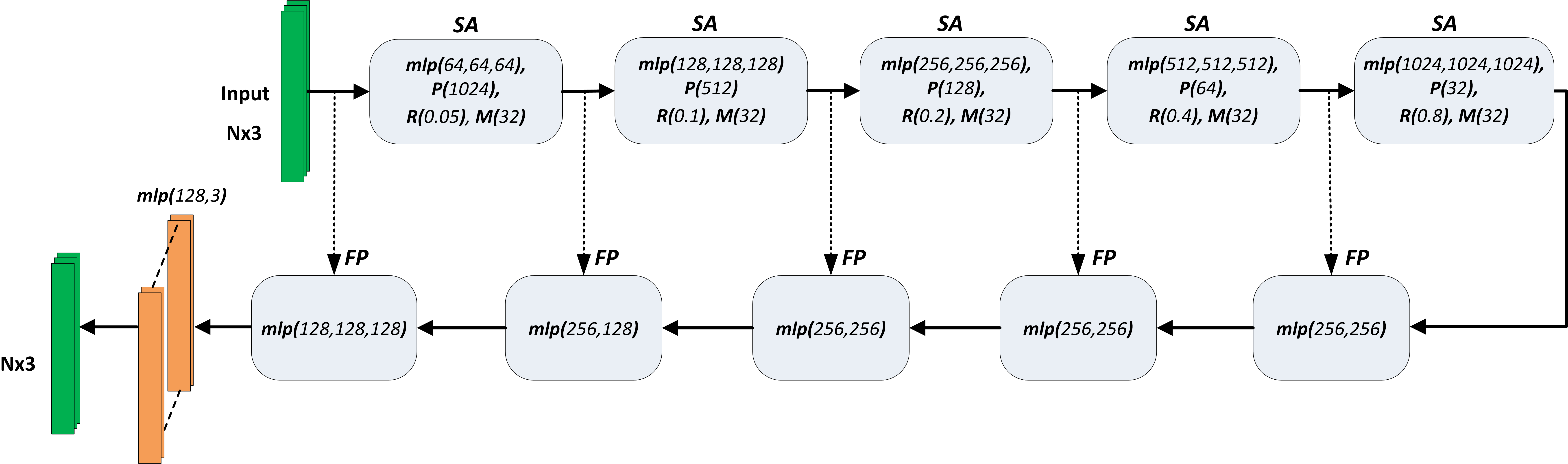}
 \caption{Modified PointNet++ architecture.}
 \label{fig:pointPlusA}
 \end{figure*}

In Fig. \ref{fig:pointPlusA}, the modified PointNet++ architecture for segmentation of rosebush point clouds is given. We increased the number of SA and FP layers from 4 to 5, adjusting the radius of the local regions ($R$) and the number of sampled points ($P$) at each layer to improve the performance on our plant models. We also altered the number of channels of MLPs within the SA and FP layers. 

\subsubsection{DGCNN}
Dynamic Graph CNN (DGCNN) architecture \cite{Wang2019} was designed to integrate local neighborhood information of 3D points directly into the network, rather than a separate grouping process as done in PointNet++. The local neighbourhood of a point is represented with a graph structure. A neural network module called EdgeConv is applied to extract edge features to encode the spatial relationship between a point and its $K$ neighbours. The edge features are extracted through MLPs applied to edge representations instead of point locations.

 \begin{figure*}[h]
 \centering
 \includegraphics[width=0.8\textwidth]{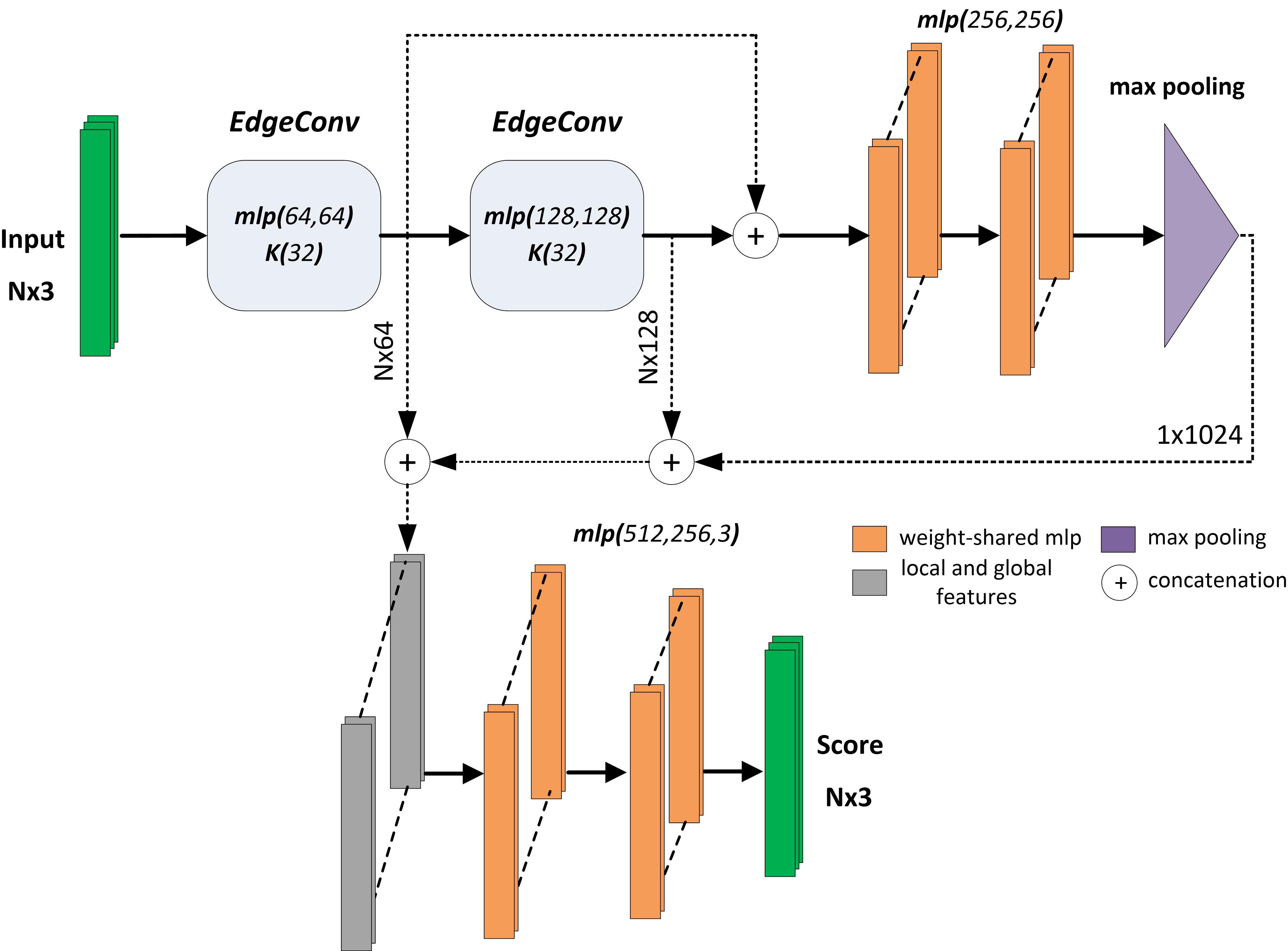}
 \caption{Modified DGCNN architecture.}
 \label{fig:DgCNNA}
 \end{figure*}

Unlike the CNN structures used in regular grids, fixed graphs are not used. The graphs are updated since the $K$ nearest neighborhoods of the point-wise features change at each layer. Only in the first layer, geometrical proximity between nearest points are considered. In the following layers, edge representations are formed between nearest neighbours that are close in the feature space. That might be an advantage in terms of diffusing the information with respect to the proximity in the feature space; however, a multi-scale hierarchical local spatial grouping is not present in DGCNN. The local geometric structure is only captured at a very localized level; i.e. only within the nearest neighbours of a point. 

The modified DGCNN architecture for segmentation is given in Fig. \ref{fig:DgCNNA}. We reduced the number of EdgeConv layers from three to two and altered the number of channels in MLPs. We increased the number of nearest neighbors $K$ used to form edge representations in spatial and feature space from 20 to 32.

\subsubsection{PointCNN}

A convolution operator that weights the features of the neighbours of a point has been introduced with PointCNN architecture \cite{Li2018}. In this convolution process defined as X-Conv, a $K \times K$-sized transformation matrix is predicted for $K$ adjacent points with multi-layer perceptrons. Typical convolution layers are then applied to the transformed features. To define larger receptive fields for convolution, representative points are generated by farthest point sampling, and features resulting from X-conv are aggregated onto these representative points. By dilating points by a factor and hierarchically applying X-conv, point features are aggregated into fewer points, representing larger spatial areas. For segmentation, point-based features are processed through an encoder-decoder structure.

 \begin{figure*}[h]
 \centering
 \includegraphics[width=1.0\textwidth]{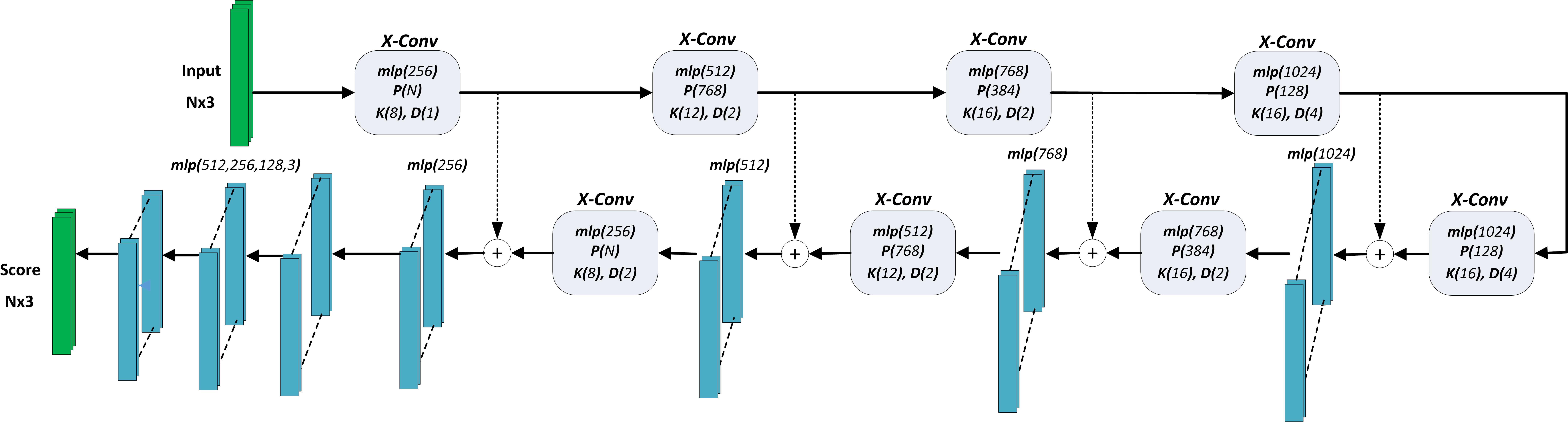}
 \caption{Modified PointCNN architecture.}
 \label{fig:PointCNNA}
 \end{figure*}

In Fig. \ref{fig:PointCNNA}, the PointCNN architecture is shown. $K$ corresponds to the number of nearest neighbours that are used in convolution. $P$ indicates the number of sampled points, and $D$ is the point dilation rate. The default values of these parameters yielded the best performance for the synthetic validation data. We inserted an additional fully connected layer and modified the number of channels in the fully connected layers prior to obtaining point-wise class scores.

\subsubsection{ShellNet}

The ShellConv convolution operator, introduced with the ShellNet architecture \cite{Zhang2019}, is applied to areas within the concentric shells of the local neighbourhood of a 3D point. The size of the sphere is increased until fixed number of points are included in each shell. Descriptive features are extracted for each shell using statistical information of the points within the shell. Since a sequence of convolution was defined outwards from starting the inner shell, the output of the convolution became relatively independent of the ordering of the points. To remove the dependency on the order of points within each shell, maximum pooling is applied to the point-wise features in the shell. ShellConv is applied hierarchically by sub-sampling the points to representative points, thus operating on larger receptive fields at subsequent layers. 

 \begin{figure*}[h]
 \centering
 \includegraphics[width=0.9\textwidth]{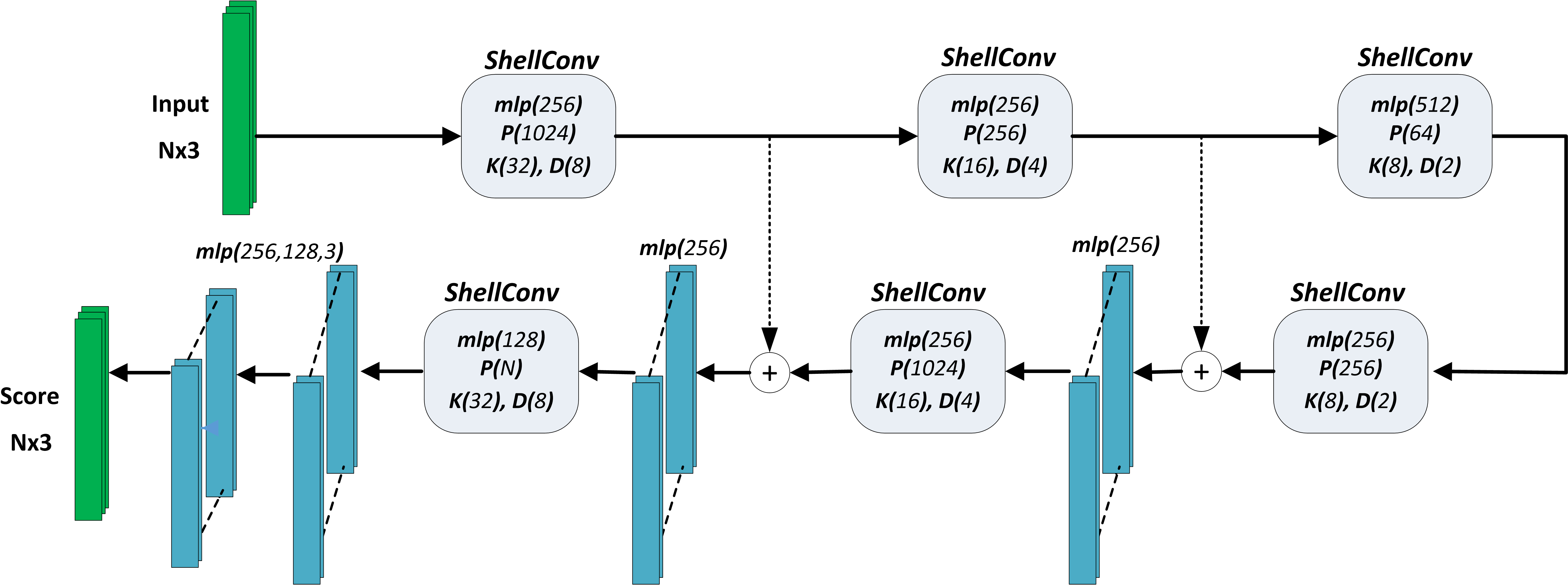}
 \caption{Modified ShellNet architecture.}
 \label{fig:shellnetA}
 \end{figure*}

 The modified ShellNet architecture for segmentation is given in Fig. \ref{fig:shellnetA}. Using the synthetic data, we tuned the parameters $P$ and $D$, corresponding to the number of sampled points in the neighborhood and the number of shells, respectively. The number of nearest neighbours ($K$) that are used in convolution was kept at its default value. We also altered the number of channels in the fully connected layers prior to obtaining point-wise class scores.

\subsubsection{RIConv}
Many 3D deep learning architectures rely on the raw 3D coordinates of the input points, hence are inherently dependent on pose variations of objects in the scene. To provide some form of rotation-invariance, data augmentation with rotated versions of the point clouds is applied. However, the networks cannot model unseen rotations. To ensure rotation invariance, a new convolution process called RIConv is proposed in \cite{Zhang2019b}. The main idea is to define the convolution process on rotation-invariant features such as angle and distance between points, rather than the raw 3D coordinates. The learned model is effective against transformations such as translation and rotation in 6-axis space. A simple binning approach for the point permutation problem is integrated into the feature extraction process. The disadvantage of aggregating distances and angles is the loss of geometric data; since two different constellations of 3D points can result in the same rotation-invariant features.

 \begin{figure*}[h]
 \centering
 \includegraphics[width=0.9\textwidth]{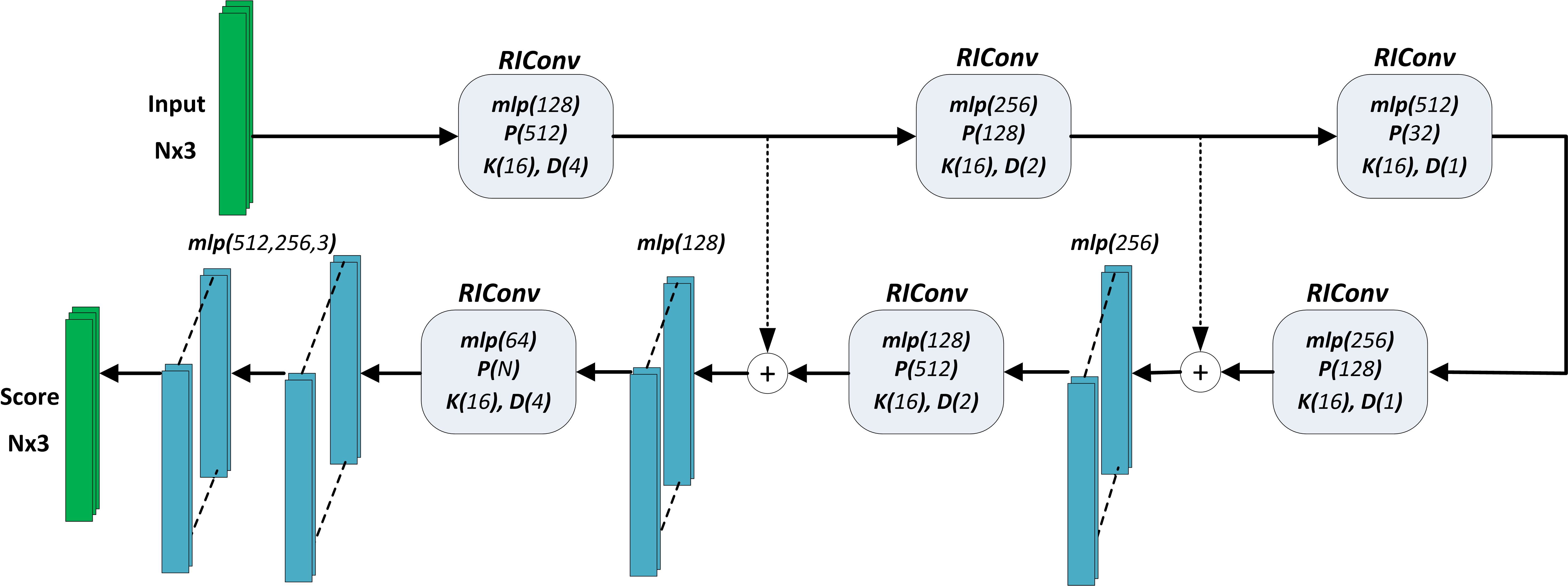}
 \caption{Modified RIConv architecture.}
 \label{fig:RIConvA}
 \end{figure*}

The encoder-decoder architectural structure of RIConv for segmentation is given in Fig. \ref{fig:RIConvA}. $K$ corresponds to the number of nearest neighbours that are used in convolution. $P$ indicates the number of sampled points, and $D$ is the number of bins. As with ShellNet, these parameters are tuned through synthetic rosebush data for RIConv, and the number of channels at the final fully-connected layers are altered for higher performance.


\section{Results}
\label{sec:results}

We adapted and tested six 3D point-based deep learning architectures for segmentation of rosebush models to their structural parts. We used recall ($Re$), precision ($Pr$) and Intersection over Union ($IoU$) to evaluate the success of each architecture. We denote the number of true positives, false positives and false negatives for each class as $TP_{C}$, $FP_{C}$, and $FN_{C}$, respectively, where $C \in \{Flower, Leaf, Stem\}$ is the class of the structural part of a rosebush. Recall ($Re$), precision ($Pr$) and Intersection over Union ($IoU$) per semantic class are then defined as
\begin{equation}
    Re = \frac{TP_C}{TP_C+FN_C}
\end{equation}

\begin{equation}
    Pr = \frac{TP_C}{TP_C+FP_C}
\end{equation}

\begin{equation}
    IoU = \frac{TP_C}{TP_C+FN_C+FP_C}\;.
\end{equation}
We also use the mean of the $IoU$ scores over all three classes ($MIoU$) and the total accuracy ($Acc$). $Acc$ is defined as the ratio of all correctly classified points to the total number of points in the model.

Using the synthetic data generated by L-studio and the real rosebush models from ROSE-X data set, we conducted seven types of experiments with each point-based deep learning algorithm:
\begin{itemize}
    
    \item \textbf{Single real rosebush model for training (I):} We used a single plant model from the ROSE-X data set of real rosebush models for training the networks. 96 blocks were extracted from the point cloud to provide training data. 20\% of the blocks were used as the validation set. The corresponding networks trained using one real rosebush plant are called as \textbf{I}-trained networks.
    \item \textbf{Two real rosebush models for training (II):} In this experiment, 159 blocks extracted from two real rosebush models are used as training data, where 20\% of the blocks are reserved for validation. The corresponding networks trained using two real rosebush models are called as \textbf{II}-trained networks.
    \item \textbf{Three real rosebush models for training (III):} In this experiment, 251 blocks extracted from three real rosebush point clouds are used as training data, where 20\% of the blocks are reserved for validation. The corresponding networks trained using three real rosebush plant are called as \textbf{III}-trained networks.
    \item \textbf{Synthetic data for training (S):} 40 of the 48 of the synthetic models generated by L-studio are used as training data. 8 models are reserved for validation. Using the results on the validation models, the parameters of each architecture are optimized. The corresponding trained networks are denoted as \textbf{S}-trained networks. 
    \item \textbf{S-trained networks updated with single real rosebush model (S+I):} The S-networks, which are initially trained and optimized with synthetic data, are re-trained using the blocks extracted from a single real rosebush model. We call these updated networks \textbf{S+I}-trained networks.
     \item \textbf{S-trained networks updated with two real rosebush models (S+II):} In this experiment, the S-networks are re-trained using the blocks extracted from two real rosebush models. We call these updated networks \textbf{S+II}-trained networks.
     \item \textbf{S-trained networks updated with three real rosebush models (S+III):} In this experiment, the S-networks are re-trained using the blocks extracted from three real rosebush models. We call these updated networks \textbf{S+III}-trained networks.
\end{itemize}

Table \ref{table:numblocks} gives the total number of training and validation blocks extracted from the synthetic and real rosebush models. Recall that the point cloud sampled from each block is treated independently by the networks. The 48 synthetic plants are partitioned such that blocks from 40 plant models are used for training and blocks from 8 plant models are used for validation. The training and validation sets of the synthetic data is used extensively to modify the networks, to determine hyper-parameters of the networks and other parameters such as block and grid sizes. For the real plant models from the ROSE-X data set, 20\% of the blocks are randomly chosen for validation from the full set of blocks reserved for training. This validation set of the real data is used to set experimentally the layers for which the weights will be updated during transfer learning \cite{Yosinski2014}. 

\begin{table}[ht]
\caption{Number of training and validation blocks used in the experiments.}
\footnotesize
 \centering
 \begin{tabular}{l  l l l} \hline 
 \textbf{ }  & \textbf{Data} & \textbf{$\#$ blocks for training} & \textbf{$\#$ blocks for validation} \\ \hline
 \textbf{S} & Synthetic & 3026 & 511 \\ \hline 
 \textbf{I} & Real &  80 & 19 \\ \hline 
 \textbf{II} & Real & 127 & 32 \\ \hline 
  \textbf{III} &  Real & 201 & 50 \\ \hline 
  \end{tabular}
  \label{table:numblocks}
 \end{table}

For the experiments where synthetic data is not involved (\textbf{I}, \textbf{II}, and \textbf{III}) the default settings of the architectures (such as number of features extracted at each layer) are left unchanged. For details of the default settings, please refer to the original articles introducing the architectures. 

For the experiments where synthetic data is used to pre-train the modified architectures (\textbf{S+I}, \textbf{S+II}, and \textbf{S+III}), the training stopped after 250 epochs. Similarly, while retraining with real data, the training stopped after 250 epochs. For all cases, the weights of the last epoch are preserved for testing. 

The hyper-parameters of the networks determined using the synthetic data are given in Table \ref{table:parameters}. 

\begin{table}[ht]
\caption{Hyper-parameters used to train the networks.}
\footnotesize
 \centering
 \begin{tabular}{l  l l l l} \hline 
 \textbf{ }  & \textbf{Learning rate} & \textbf{Batch} & \textbf{Decay step/rate}& \textbf{Weight decay} \\ \hline
 \textbf{PointNet} &0.001&48&30000/0.8&0.005\\ \hline 
 \textbf{PointNet++} &0.005&12&200000/0.7&None\\ \hline 
 \textbf{DGCNN} &0.005&12&200000/0.5&None\\ \hline 
  \textbf{PointCNN} &0.005&8&10000/0.8&1e-8\\ \hline 
 \textbf{ShellNet} &0.005&12&5000/0.8&1e-8\\ \hline 
 \textbf{RIConv}   &0.005&12&10000/0.8&1e-6\\ \hline 
 \end{tabular}
  \label{table:parameters}
 \end{table}

Table \ref{table:LStudioResult} gives the segmentation results of the \textbf{S}-trained networks on the 8 synthetic validation models. PointNet++, DGCNN, ShellNet and PointCNN were able to produce performance success over 90\% for all measures. For the synthetic models, local geometric variations at the organ level (e.g. leaf shape, branch thickness) are limited to the variations imposed by the generation rules of the simulator. Hence, the networks were easily able to model the geometric characteristics that distinguish the three organs.  PointNet produced an $MIoU$ below 60\% due to its inability to encode geometric information at various scales.

For the rest of the experiments, the networks are tested on the point clouds extracted from 8 real rosebush models from the ROSE-X data set through block partitioning. The predictions on the blocks are merged to obtain the final segmentation of the full plant models as described in the section for data preprocessing.

\begin{figure}[H]
	\centering
	\includegraphics[width=0.75\textwidth]{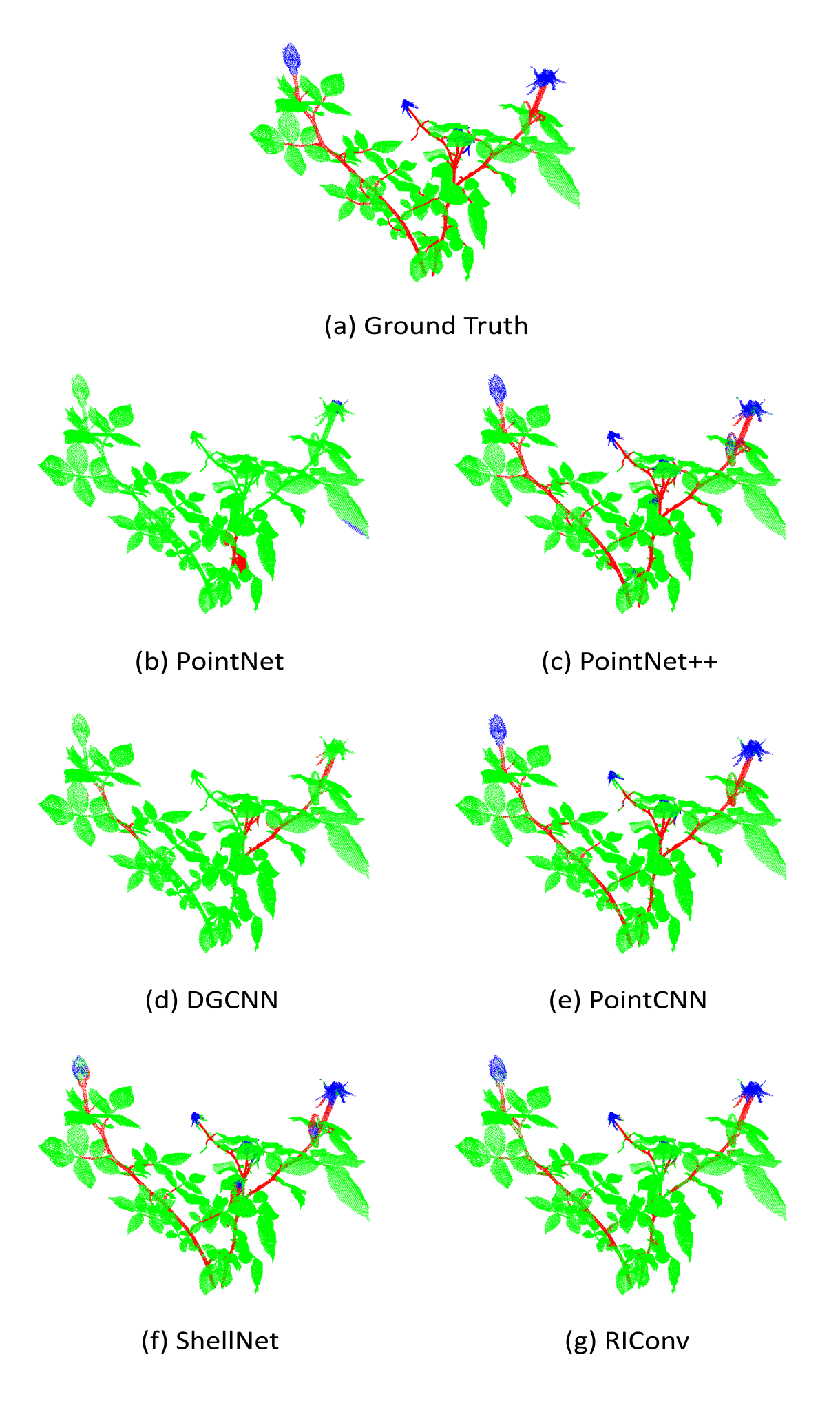}
	\caption{A real rosebush model segmented with the networks trained with with three real rosebush models (\textbf{III}).}
    \label{fig:sampleSegmented_III}
\end{figure}

\begin{figure}[H]
	\centering
    \includegraphics[width=0.75\textwidth]{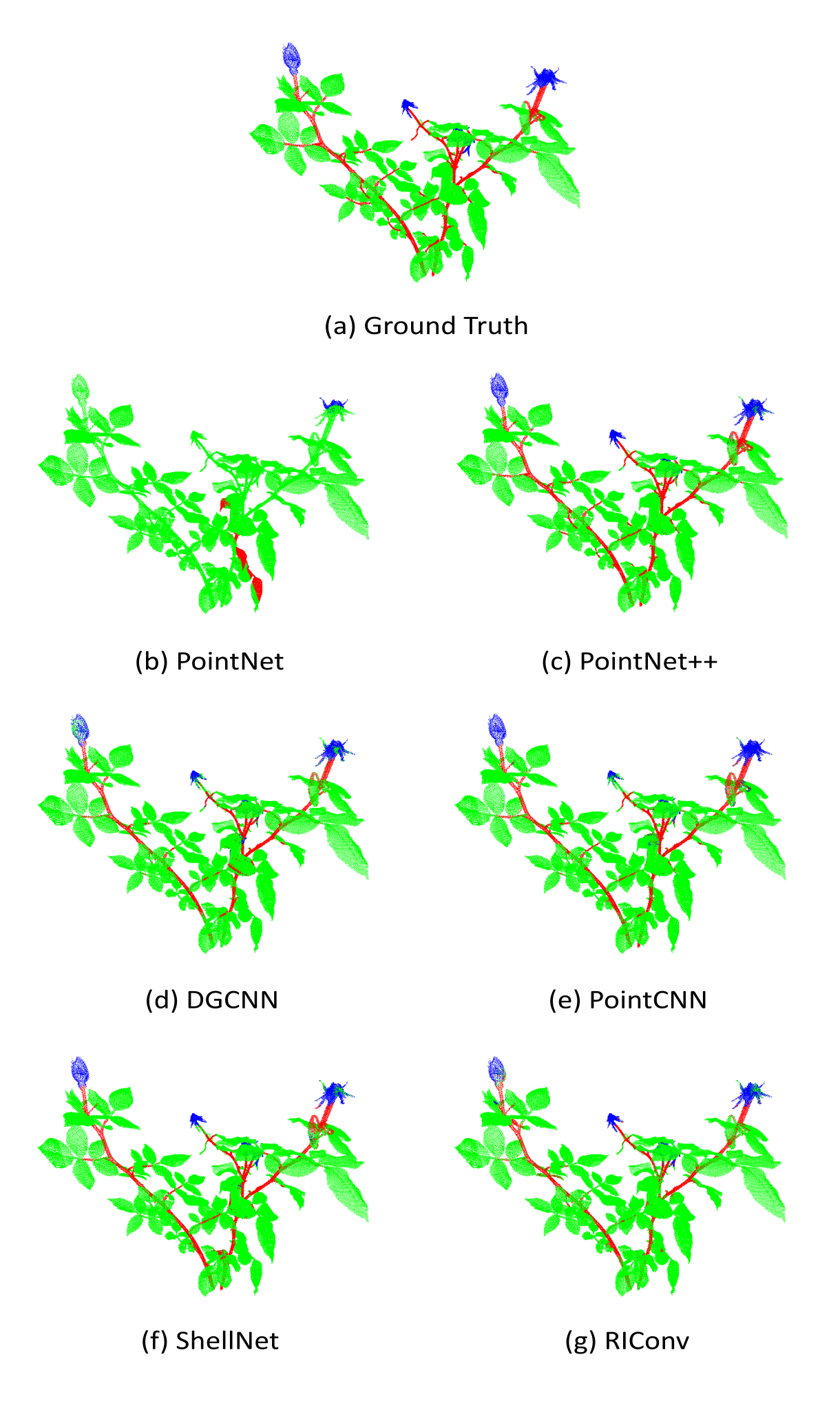}
	\caption{A real rosebush model segmented with the networks trained with synthetic models and updated with three real rosebush models (\textbf{S+III}).}
    \label{fig:sampleSegmented_S_III}
\end{figure}

In Fig. \ref{fig:sampleSegmented_III}, we visualized the segmentation results on a sample real rosebush model obtained with \textbf{III}-trained networks; i.e. only three real rosebush models were used for training. In Fig. \ref{fig:sampleSegmented_S_III}, the segmentation results on the same test model with \textbf{S+III}-trained networks are given.

 \begin{table}[ht]
 \caption{Segmentation results on the validation set of the 8 synthetic rosebush models. 40 synthetic rosebush models were used to train the networks.}
 \footnotesize
 \centering
 \begin{tabular}{l l p{1.1cm} p{1.6cm} p{1.1cm} p{1.1cm} p{1.3cm} p{1.1cm} }
 \hline 
 \multicolumn{2}{l}{} & \textbf{PointNet} &  \textbf{PointNet++} & \textbf{DGCNN} & \textbf{ShellNet} & \textbf{PointCNN}& \textbf{RIConv}\\ \hline
 \multirow{3}{*}{$Re$} &Flower &64.85&99.48&99.21&99.64&99.71&97.29\\&Leaf &95,28&98.43&99.59&99.54&99.56&97.98 \\&Stem &45.16&97.82&97.36&96.87&98.23&78.71 \\ \hline 
 \multirow{3}{*}{$Pr$} &Flower &88.19&99.78&99.68&99.59&99.59&98.57 \\&Leaf &81.42&99.39&99.28&99.20&99.56&94.28\\&Stem &77.39&94.04&98.15&98.22&98.32&91.60  \\ \hline
 \multirow{3}{*}{$IoU$} &Flower &59.67&99.26&98.90&99.24&99.30&95.94 \\&Leaf &78.27&97.84&98.87&98.74&99.13&92.48\\&Stem &39.90&92.11&95.60&95.19&96.61&73.41\\ \hline
 \multicolumn{2}{l}{$MIoU$} &59.28&96.40&97.79&97.73&98.35&87.28\\ \hline
 \multicolumn{2}{l}{$Acc$} &81.82&98.50&99.15&99.10&99.36&94.57\\
 \hline
 \end{tabular}
  \label{table:LStudioResult}
 \end{table}

Table \ref{table:pointNetResult} gives the segmentation results obtained with PointNet on the real test plants. Columns in Table \ref{table:pointNetResult} correspond to the segmentation results of the seven types of experiments. The results correspond to the performance values averaged over 8 models. Despite the increase in the training data and the incorporation of synthetic data, the segmentation performance of PointNet is low, especially for the flower and stem parts. Not being able to capture the distinguishing geometrical structures of the parts, PointNet seems to favor the leaf class due to the imbalance in the training data (Fig. \ref{fig:sampleSegmented_III}b). 

\begin{table}[ht]
\caption{Segmentation results on 8 real rosebush models from ROSE-X data set with PointNet.}
\footnotesize
 \centering
 \begin{tabular}{p{0.7cm} p{0.8cm} C{0.8cm} C{0.8cm} C{0.8cm} C{0.8cm} C{1.1cm} C{1.1cm} C{1.1cm}}
 \hline 
 \multicolumn{2}{l}{\textbf{PointNet}} & \textbf{I} & \textbf{II} & \textbf{III} & \textbf{S} & \textbf{S + I} & \textbf{S + II} & \textbf{S + III}\\ \hline
 \multirow{3}{*}{$Re$} &Flower  &10.50&11.92&19.18&20.44&14.79&13.85&8.45 \\ &Leaf  &94.77&97.77&97.55&96.52&90.43&94.63&96.18 \\& Stem &5.47&2.30&3.15&2.20&9.21&7.25&8.62 \\ \hline
 \multirow{3}{*}{$Pr$} &Flower &27,43&34.67&40.35&41.61&31.48&34.95&41.24\\ &Leaf  &81.77&82.05&82.62& 82.52&82.18&82.52&82.06\\&Stem &19.33&29.13&45.07&15.9&14.41&20.16&28.07 \\ \hline
 \multirow{3}{*}{$IoU$} &Flower &8.22&9.73&14.94&15.88&11.19&11.01&7.54\\ &Leaf  &78.23&80.54& 80.94&80.14&75.60&78.83&79.48\\&Stem &4.46&2.18&3.03&1.99&5.95&5.63&7.06\\ \hline
 \multicolumn{2}{l}{$MIoU$}  &30.30&30.82&32.97&32.66&30.92&31.83&31.36\\ \hline
 \multicolumn{2}{l}{$Acc$}  &78.17&80.35&80.81&79.96&75.41&78.52&79.53\\
 \hline
 \end{tabular}
 \label{table:pointNetResult}
 \end{table}

\begin{figure}[ht]
 	\centering
	\includegraphics[width=0.7\textwidth]{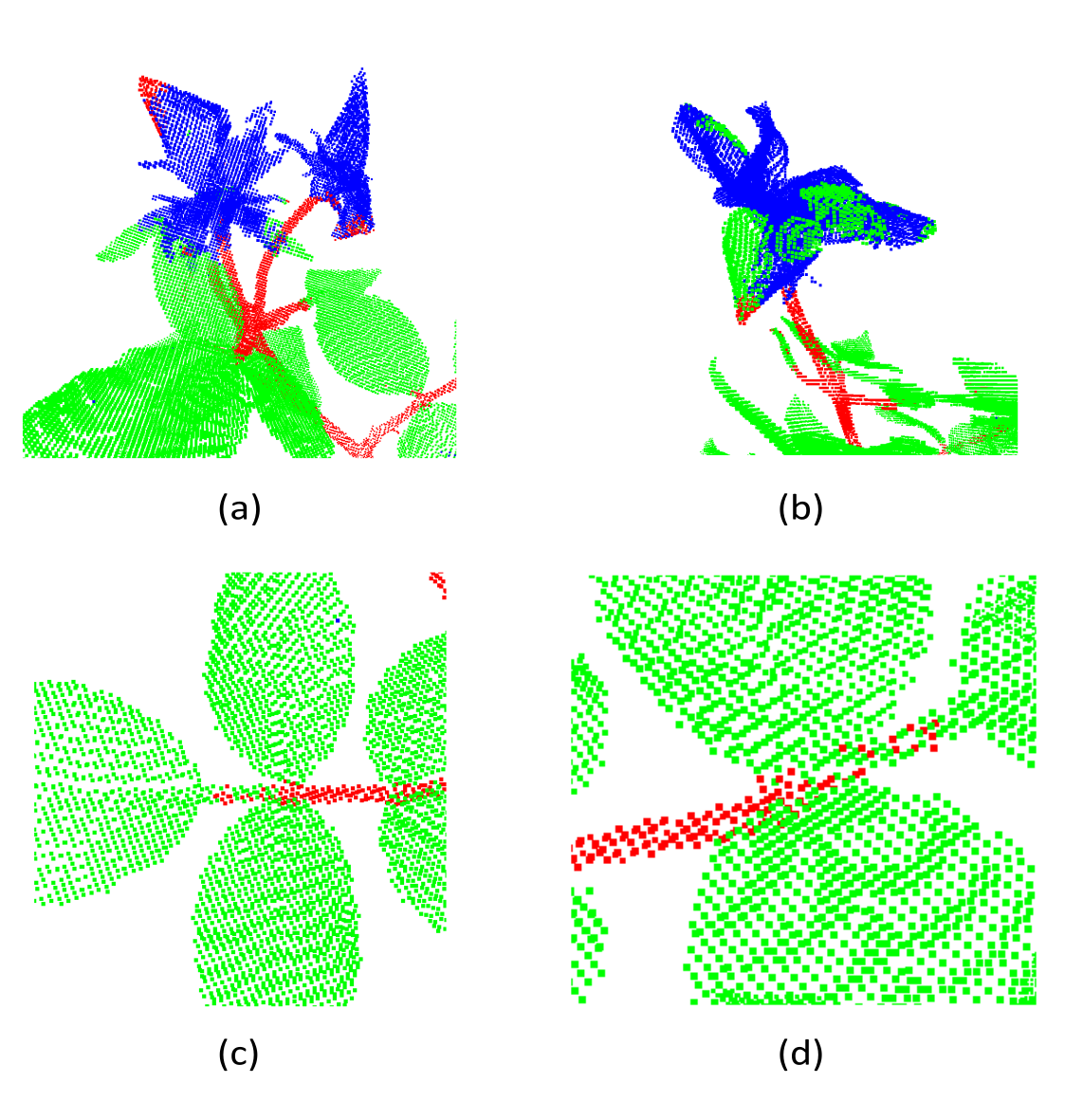}
	\caption{Examples to erroneous segmentation results produced by PointNet++ (S+III).}
    \label{fig:errpointP}
\end{figure}

The segmentation results of 8 test real rosebush models yielded by PointNet++ with seven experimental setups are given in Table \ref{table:pointNet++Result}. The increase of the training data from a single rosebush model to two and then three models led to an increase in the performance, especially for the stem class. The use of synthetic data alone for training was not effective; however when the network pre-trained with synthetic data was updated with real rosebush models the performance was improved. The results with PointNet++ are promising with an accuracy rate over 95\% and a mean $IoU$ rate over 85\%. The main sources of errors are the confusion between stems and thick parts of flowers (Fig. \ref{fig:errpointP}a), between leaves and petals of flowers (Fig. \ref{fig:errpointP}b), and between petioles and leaves (Fig. \ref{fig:errpointP}c, \ref{fig:errpointP}d).

 \begin{table}[ht]
 \caption{Segmentation results on 8 real rosebush models from ROSE-X data set with PointNet++.}
 \footnotesize
 \centering
 \begin{tabular}{p{0.7cm} p{0.8cm} C{0.8cm} C{0.8cm} C{0.8cm} C{0.8cm} C{1.1cm} C{1.1cm} C{1.1cm}}
 \hline 
 \multicolumn{2}{l}{\textbf{PointNet++}} & \textbf{I} & \textbf{II} & \textbf{III} & \textbf{S} & \textbf{S + I} & \textbf{S + II} & \textbf{S + III}\\ \hline
 \multirow{3}{*}{$Re$} & Flower &69.98&68.79&81.24&97.02&75.13&73.70&85.39 \\ &Leaf &98.02&98.28&97.67&28.72&98.10&98.71&98.71 \\&Stem &78.73&85.75&87.55&48.81&84.57&87.73&89.03 \\ \hline 
 \multirow{3}{*}{$Pr$} &Flower &87.80&95.10&87.30&10.60&84.26&91.61&91.57 \\ &Leaf &95.42&96.12&97.27&96.01&96.39&96.55&97.58\\&Stem
 &83.73&83.89&86.20&78.09&89.47&90.48&92.52 \\ \hline
 \multirow{3}{*}{$IoU$} &Flower &63.78&66.44&72.72&10.57&65.88&69.04&79.17 \\ &Leaf &93.62&94.53&95.07&28.38&94.62&95.35&96.36\\&Stem
 &68.29&73.63&76.79&42.93&76.91&80.32&83.05 \\ \hline
 \multicolumn{2}{l}{$MIoU$} &75.23&78.20&81.53&27.29&79.14&81.57&86.19\\ \hline
 \multicolumn{2}{l}{$Acc$} &93.70&94.63&95.28&36.15&94.82&95.58&96.60\\
 \hline
 \end{tabular}
 \label{table:pointNet++Result}
 \end{table}

The effect of using synthetic data on the segmentation results is even more pronounced for DGCNN (Table \ref{table:DgCNNResult}), PointCNN (Table \ref{table:PointCNNResult}), and ShellNet (Table \ref{table:ShellNetResult}). Rather than training a network with real data from scratch (as in the cases of \textbf{I}, \textbf{II}, and \textbf{III}), using the real data to fine-tune a network trained by synthetic data (as in the cases of \textbf{S+I}, \textbf{S+II}, and \textbf{S+III}) boosts the performance, especially for the stem and flower classes.

 We can observe from Fig. \ref{fig:sampleSegmented_III}d that with DGCNN, parts of main stems were classified as leaves and the flower class is not retrieved at all (27.94\% and 7.12\% recall rates for the flower and stem classes, respectively, in Table \ref{table:DgCNNResult}). We conjecture that DGCNN is only encoding the geometric structure at the very local level; the spatial receptive field was limited to the K-neighbours of each point in 3D. The data imbalance in the training data in favor of leaves limited the capacity of DGCNN to learn features from stem and flower regions. The effect of data imbalance was alleviated with incorporating synthetic data in training data as seen in Fig. \ref{fig:sampleSegmented_S_III}d. DGCNN was able to capture branch and flower structures with pre-training with synthetic models.

\begin{figure}[ht]
 	\centering
	\includegraphics[width=0.7\textwidth]{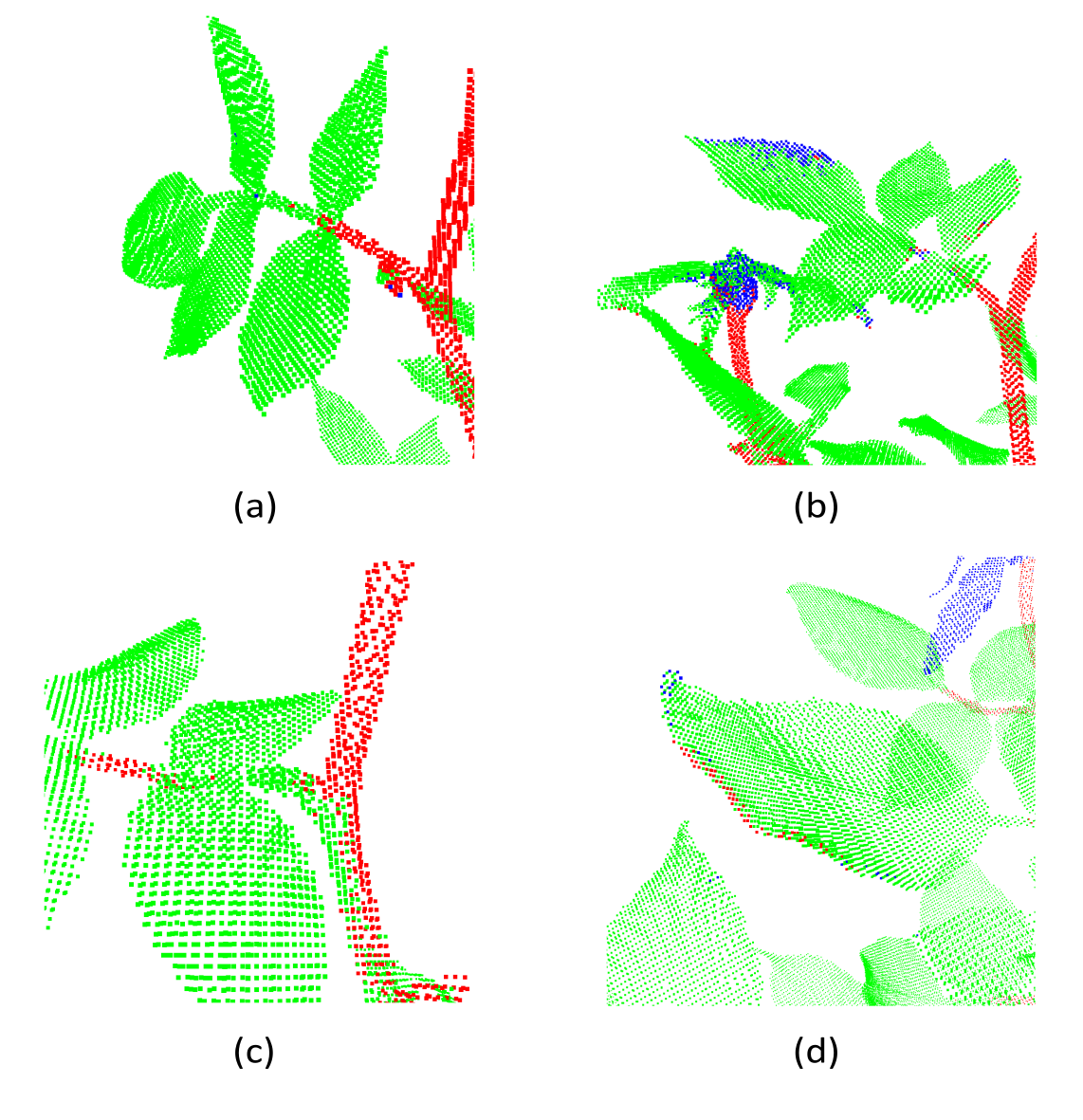}
	\caption{Examples to erroneous segmentation results produced by DGCNN (S+III).}
    \label{fig:errDgCNN}
\end{figure}

Despite the incorporation of synthetic data, DGCNN's performance lacks behind PointNet++, PointCNN, and ShellNet. These three architectures, in contrast to DGCNN, have the capacity to increase the size of the spatial receptive fields through successive re-grouping and feature aggregation. Examples to erroneous segmentation results produced by DGCNN are visualized in Fig. \ref{fig:errDgCNN}. Classifying petioles as leaves (Fig. \ref{fig:errDgCNN}a) is a common error for all architectures, however it occurs more frequently with DGCNN. Confusion between leaves and flowers are present (Fig. \ref{fig:errDgCNN}b). Surfaces of main stems can be classified as leaf points (Fig. \ref{fig:errDgCNN}c). In some cases, boundaries of leaves are assigned to the stem class (Fig. \ref{fig:errDgCNN}d).

 \begin{table}[ht]
 \caption{Segmentation results on 8 real rosebush models from ROSE-X data set with DGCNN.}
  \footnotesize
 \centering
 \begin{tabular}{p{0.7cm} p{0.8cm} C{0.8cm} C{0.8cm} C{0.8cm} C{0.8cm} C{1.1cm} C{1.1cm} C{1.1cm}}
 \hline 
 \multicolumn{2}{l}{\textbf{DGCNN}} & \textbf{I} & \textbf{II} & \textbf{III} & \textbf{S} & \textbf{S + I} & \textbf{S + II} & \textbf{S + III}\\ \hline
 \multirow{3}{*}{$Re$} & Flower &5.62&29.85&7.12&79.65&24.36&34.83&59.16\\ &Leaf &97.85&95.93&98.58&84.27&95.91&97.45&98.20\\&Stem
 &9.68&28.59&27.94&49.05&46.81&73.44&67.35\\ \hline 
 \multirow{3}{*}{$Pr$} &Flower &22.67&54.80&75.96&42.46&37.51&75.46&79.71\\ &Leaf &82.34&86.80&84.60&92.26&88.05&92.71&93.27\\&Stem
 &58.97&51.36&65.42&47.21&77.97&74.16&84.65\\ \hline
 \multirow{3}{*}{$IoU$} &Flower &4.72&23.95&6.96&38.30&17.33&31.28&51.42\\ &Leaf &80.88&83.72&83.57&78.71&84.87&90.51&91.71\\&Stem
 &9.07&22.50&24.34&31.68&41.35&58.48&60.02\\ \hline
 \multicolumn{2}{l}{$MIoU$} &31.56&43.39&38.29 &49.56&47.85&60.09&67.72\\ \hline
 \multicolumn{2}{l}{$Acc$} &80.79&83.23&83.59&79.87&84.90&90.00&91.73\\
 \hline
 \end{tabular}
 \label{table:DgCNNResult}
 \end{table} 
 
\begin{figure}[ht]
 	\centering
	\includegraphics[width=0.7\textwidth]{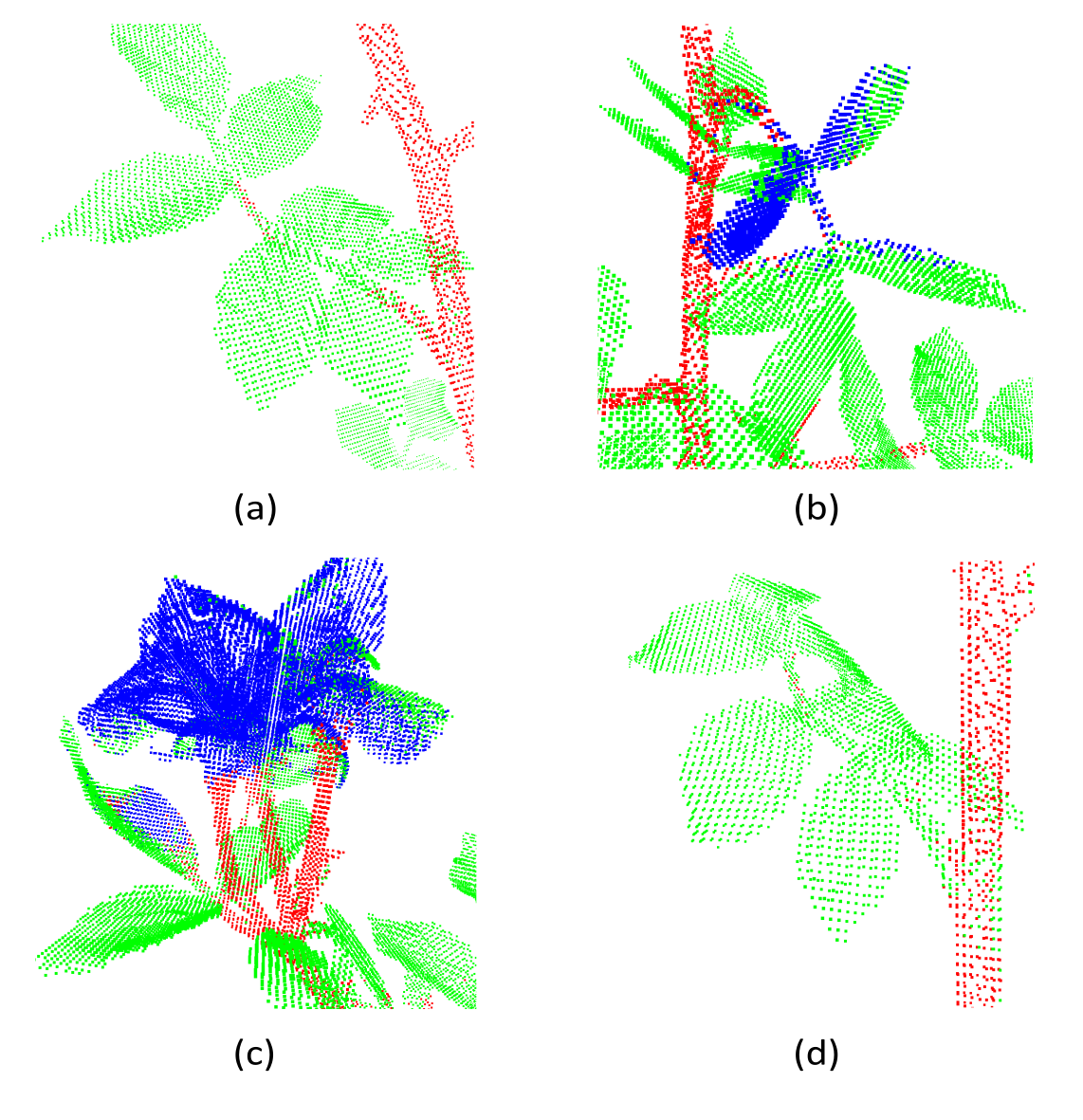}
	\caption{Examples to erroneous segmentation results produced by PointCNN (S+III).}
    \label{fig:errpointcnn}
\end{figure}

The second best results after PointNet++ were obtained with PointCNN (Table \ref{table:PointCNNResult}). Examples to erronous segmentation results produced by PointCNN are shown in Fig. \ref{fig:errpointcnn}. We observe petioles classified as leaves (Fig. \ref{fig:errpointcnn}a and \ref{fig:errpointcnn}d), and elongated and thick leaves classified as flowers (Fig. \ref{fig:errpointcnn}b). There is also confusion between leaves and petals (Fig. \ref{fig:errpointcnn}c). In some cases, main stem points close to leaves are classified as leaf points (Fig. \ref{fig:errpointcnn}d).

 \begin{table}[ht]
  \caption{Segmentation results on 8 real rosebush models from ROSE-X data set with PointCNN.}
 \footnotesize
 \centering
 \begin{tabular}{p{0.7cm} p{0.8cm} C{0.8cm} C{0.8cm} C{0.8cm} C{0.8cm} C{1.1cm} C{1.1cm} C{1.1cm}}
 \hline 
 \multicolumn{2}{l}{\textbf{PointCNN}} & \textbf{I} & \textbf{II} & \textbf{III} & \textbf{S} & \textbf{S + I} & \textbf{S + II} & \textbf{S + III}\\ \hline
 \multirow{3}{*}{$Re$} &Flower &5.40&23.49&52.55&97.04&72.90&75.52&68.21\\&Leaf &99.18&99.00&99.18&59.04&97.75&97.86&98.41\\&Stem &34.06&66.46&72.39&46.53&48.08&72.79&75.77\\ \hline 
 \multirow{3}{*}{$Pr$} &Flower &80.53&92.07&90.32&19.94&75.83&79.56&84.35\\&Leaf &84.85&89.85&92.87&95.81&91.80&95.06&94.83\\&Stem
 &81.31&86.55&91.21&39.33&84.56&88.14&88.23\\ \hline
 \multirow{3}{*}{$IoU$} &Flower &5.33&23.03&49.76&19.82&59.15&63.25&60.55\\&Leaf &84.25&89.04&92.16&57.55&89.90&93.12&93.40\\&Stem &31.58&60.24&67.67&27.09&44.20&66.30&68.81\\ \hline
 \multicolumn{2}{l}{$MIoU$} &40.39&57.44&69.86&34.82&64.42&74.22&74.26\\ \hline
 \multicolumn{2}{l}{$Acc$} &84.65&89.60&92.60&60.45&90.17&93.30&93.54\\
 \hline
 \end{tabular}
 \label{table:PointCNNResult}
 \end{table}

The quantitative performance results obtained with ShellNet architecture (Table \ref{table:ShellNetResult}) are close to those of PointCNN. They use similar strategies to group local points; they both recursively sub-sample the point cloud through selecting representative points and aggregate features from the closest neighbours of these representatives. In PointCNN, however, aggregation through convolution is performed through a predicted ordering of all the neighbour points; a property to which we attribute its higher performance compared to ShellNet.

\begin{figure}[ht]
	\centering
	\includegraphics[width=0.7\textwidth]{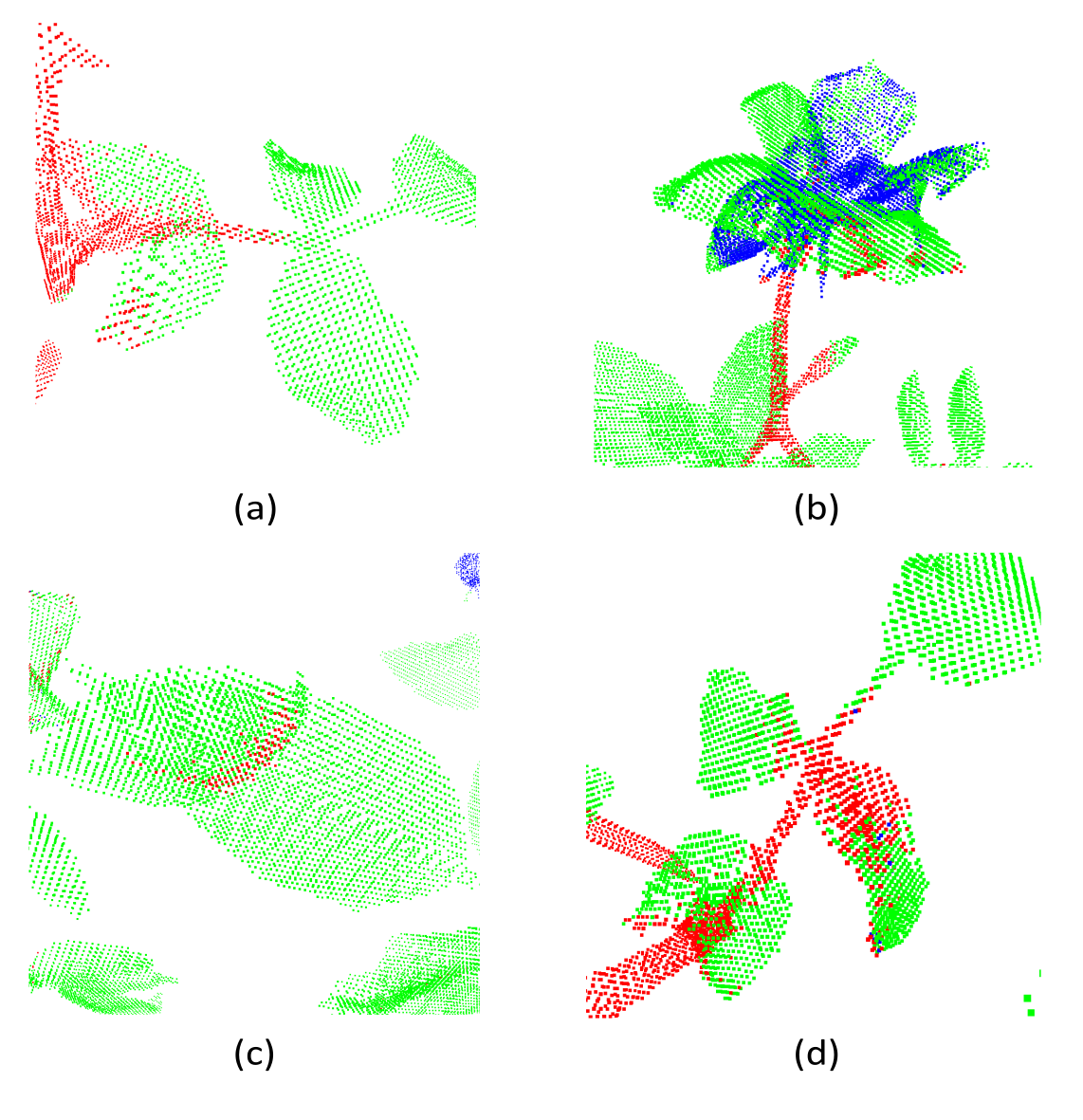}
	\caption{Examples to erroneous segmentation results produced by ShellNet (S+III).}
    \label{fig:errshellnet}
\end{figure}

With ShellNet, as with the other architectures, petioles (Fig. \ref{fig:errshellnet}a) and petals (Fig. \ref{fig:errshellnet}b) were occasionally confused with leaf points. Touching leaves resulting in thick structures are also a cause of error (Fig. \ref{fig:errshellnet}c). Another source of error with ShellNet is the interference of points from close parts, such as the misclassifications of leaf points as stems (Fig. \ref{fig:errshellnet}d).  

 \begin{table}[ht]
  \caption{Segmentation results on 8 real rosebush models from ROSE-X data set with ShellNet.}
 \footnotesize
 \centering
\begin{tabular}{p{0.7cm} p{0.8cm} C{0.8cm} C{0.8cm} C{0.8cm} C{0.8cm} C{1.1cm} C{1.1cm} C{1.1cm}}
 \hline 
 \multicolumn{2}{l}{\textbf{ShellNet}} & \textbf{I} & \textbf{II} & \textbf{III} & \textbf{S} & \textbf{S + I} & \textbf{S + II} & \textbf{S + III}\\ \hline
 \multirow{3}{*}{$Re$}&Flower &48.62&44.32&51.54&97.27&68.13&51.12&58.95\\&Leaf &96.52&97.39&98.09&66.19&95.74&98.62&98.62\\&Stem
 &30.83&54.11&59.87&41.85&58.17&66.79&73.48\\ \hline 
 \multirow{3}{*}{$Pr$}&Flower &64.54&82.61&85.05&20.06&74.14&87.63&91.26\\&Leaf &87.85&90.35&91.46&94.40&92.59&92.24&93.38\\&Stem
 &66.20&71.80&80.24&70.06&71.54&85.15&87.90\\ \hline
 \multirow{3}{*}{$IoU$}&Flower &38.37&40.54&47.26&19.95&55.05&47.67&55.80\\&Leaf &85.16&88.22&89.86&63.69&88.93&91.07&92.18\\&Stem
 &26.64&44.63&52.18&35.50&47.24&59.82&66.73\\ \hline
 \multicolumn{2}{l}{$MIoU$} &50.05&57.80&63.10&39.71&63.74&69.19&71.57\\ \hline
 \multicolumn{2}{l}{$Acc$} &85.37&88.43&90.20&65.72&89.35&91.40&92.75\\
 \hline
 \end{tabular}
 \label{table:ShellNetResult}
 \end{table}

The segmentation results obtained with RIConv (Table \ref{table:RIConvResult}) fall behind all the architectures except PointNet. The local regions were extracted in the same way as in ShellNet, however, use of rotation invariant features resulted in significant loss of geometric information about the constellation of the points, which is especially important in distinguishing plant parts.  

 \begin{table}[ht]
  \caption{Segmentation results on 8 real rosebush models from ROSE-X data set with RIConv.}
 \footnotesize
 \centering
\begin{tabular}{p{0.7cm} p{0.8cm} C{0.8cm} C{0.8cm} C{0.8cm} C{0.8cm} C{1.1cm} C{1.1cm} C{1.1cm}}
 \hline 
 \multicolumn{2}{l}{\textbf{RIConv}} & \textbf{I} & \textbf{II} & \textbf{III} & \textbf{S} & \textbf{S + I} & \textbf{S + II} & \textbf{S + III}\\ \hline
 \multirow{3}{*}{$Re$} &Flower &45.99&40.40&56.09&81.47&38.16&61.00&55.47\\&Leaf &98.46&98.36&98.92&72.85&99.27&98.26&98.48\\&Stem
 &31.53&38.09&37.88&38.58&40.62&55.67&57.88\\ \hline 
 \multirow{3}{*}{$Pr$} &Flower &85.76&88.46&91.81&21.21&90.16&89.61&92.04\\&Leaf &87.49&87.82&89.08&90.17&87.87&91.45&91.33\\&Stem
 &75.03&74.86&79.61&76.89&87.69&81.41&81.78\\ \hline
 \multirow{3}{*}{$IoU$} &Flower &42.72&38.38&53.42&20.24&36.63&56.97&52.93\\&Leaf &86.31&86.55&88.22&67.49&87.31&90.00&90.07\\&Stem
 &28.54&33.77&34.53&34.58&38.43&49.39&51.27\\ \hline
 \multicolumn{2}{l}{$MIoU$} &52.52&52.90&58.72&40.77&54.12&65.45&64.76\\ \hline
 \multicolumn{2}{l}{$Acc$} &86.82&87.08&88.68&69.55&87.94&90.57&90.59\\
 \hline
 \end{tabular}
 \label{table:RIConvResult}
 \end{table}

All networks, with the exception of PointNet, when trained with synthetic data only, yield relatively high recall and low precision for the flower class on real rosebush plants. We conjecture that the reason is the mismatch of the flower class betweeen synthetic data and real plants in terms of both geometrical structure and the ratio of occurrence. High recall together with low precision for the flower class means that the networks are biased towards classifying a significant portion of leaves as flowers, causing low recall values for the leaves. When the networks are updated with real training plants, this bias is compensated and the precision for the flower class and the recall for the leaf class improve.

In general, the $mIoU$ increases as the networks are updated with more real training data. However, for PointCNN (Table \ref{table:PointCNNResult}), the improvement between the cases S+II and S+III is not significant, and for RIConv (Table \ref{table:RIConvResult}) $MIoU$ drops about 1\% with S+III compared to S+II. For both networks, the recall for the flower class decreases as the number of real training plants is increased from two to three. More petioles are classified as leaves, as these two networks start to favor classifying elongated structures as leaves, which in turn translates into a drop in the precision of leaves. Despite this observation, PointCNN gives the second best $IoU$ for the flower class among all the networks for the case S+III (Table \ref{table:comparedResult}).

\begin{table}[ht]
 \caption{Segmentation results on 8 real rosebush models for all architectures. The first row for each class corresponds to $IoU$ results of networks trained with three real rosebush models (III). The second row for each class gives the $IoU$ results for the case, where the networks were trained with synthetic models and updated with three real rosebush models (S+III). The third row for each class gives the gain in $IoU$ obtained by incorporating synthetic models. The last three rows of the Table corresponds to $MIoU$ over all classes.}
\footnotesize
 \centering
 \begin{tabular}{l l l l l l l l l l l l l l l }
 \hline 
 \multicolumn{2}{l}{} & \textbf{PointNet} & \textbf{PointNet++} & \textbf{DGCNN} & \textbf{PointCNN} & \textbf{ShellNet} & \textbf{RIConv}\\ \hline
 \multirow{3}{*}{Flower} &III &14.94&72.72&6.96&49.76&47.26&53.42\\&S+III &7.54&79.17&51.42&60.55&55.80&52.93\\&Gain
 &\textbf{-7.40}&\textbf{+6.45}&\textbf{+44.46}&\textbf{+10.79}&\textbf{+8.54}&\textbf{-0.49}\\ \hline 
 \multirow{3}{*}{Leaf} &III &80.94&95.07&83.57&92.16&89.86&88.22\\&S+III &79.48&96.36&91.71&93.40&92.18&90.07\\&Gain
 &\textbf{-1.46}&\textbf{+1.29}&\textbf{+8.14}&\textbf{+1.24}&\textbf{+2.32}&\textbf{+1.85}\\ \hline 
 \multirow{3}{*}{Stem} &III &3.03&76.79&24.34&67.67&52.18&34.53\\&S+III &7.06&83.05&60.02&68.81&66.73&51.27\\&Gain
 &\textbf{+4.03}&\textbf{+6.26}&\textbf{+35.68}&\textbf{+1.14}&\textbf{+14.55}&\textbf{+16.74}\\ \hline 
 \multirow{3}{*}{$MIoU$} &III &32.97&81.53&38.29&69.86&63.10&58.72\\&S+III &31.36&86.19&67.72&74.26&71.57&64.76\\&Gain
 &\textbf{-1.61}&\textbf{+4.66}&\textbf{+29,43}&\textbf{+4.40}&\textbf{+8.47}&\textbf{+6.04}\\ \hline 
 \end{tabular}
 \label{table:comparedResult}
 \end{table}

To summarize the results and to demonstrate the effect of incorporation of synthetic models, we give the segmentation performances of all architectures with \textbf{III}-trained and \textbf{S+III}-trained networks in Table \ref{table:comparedResult}. The use of synthetic data was beneficial for almost all classes and all architectures, except for PointNet. There is a slight decrease in the $IoU$ value for the flower class with RIConv, which is compensated by a significant increase in the performance for the stem class.

We can also observe from Table \ref{table:comparedResult} that RIConv performed poorly compared to other architectures due to the information loss with rotation invariant features. DGCNN used a single spatial receptive field at the very local level and opted for feature proximity in a non-local way; therefore missing the multi-scale spatial variability in plant parts. 

The best results were obtained with PointNet++ with or without the use of synthetic data for training. The hierarchically organized local regions for feature extraction with PointNet++ are defined in terms of metric radius. The spatial hierarchy is flexible and can be adjusted without changing the network structure. The next best two methods are PointCNN and ShellNet, both of which hierarchically regroup points and aggregate features within the network. However, the neighbourhoods are defined with respect to K-neighbourhood of points instead of metric radius. Therefore, it is not straightforward to adjust the size of the receptive fields for these architectures while taking into account both the size of the plant structures and the point density of the point clouds.

\section{Discussion}

In their default settings, the design parameters (such as number of features and layers) of the six networks and other hyperparameters (such as the radii of local regions) were originally adjusted for 3D datasets which contain point cloud scenes of indoor environments and cityscapes. The general practice for adjusting such parameters is to search for the best-performing settings through experimentation with a validation set. In our case, since we have limited data for real rosebush models, we used a subset of the synthetic dataset as validation set, systematically varied the design parameters without altering the general structure and modified each network so as to maximize its performance on the validation set. The objective was to provide a fair comparison among the six networks, whose default parameters were determined using data domains  different from plant data.

Methodological research is ongoing to automatically adjust not only the hyperparameters but the entire architecture of the network \cite{Liu2019auto}. So far, the effectiveness of genetic algorithms for the search of design parameters was demonstrated with convolutional networks \cite{wei2021genetic}. This could stand as an interesting perspective to explore such approaches with point cloud based neural networks.

While designing a 3D point-based architecture to operate effectively on plant data, an important consideration is the multi-scale and self-similar nature of plants. The architecture should be able to handle multiple, hierarchical spatial receptive fields in the network and their sizes should be easily tuned to the scales of various structures in the plants. The multi-scale feature extraction scheme is also necessary to account for the intra-class size variations; such as variations in branch diameter or leaf length and intra-class geometric variations, such as diverse range of curvature on the branches and leaves. Also grouping features with respect to their proximity in the feature space can lead to non-local similarity modeling to capture repetitive structures that are inherent to plants. 

The robustness of the architecture to heterogeneous point density, missing information and reconstruction noise is an important factor, especially for 3D data obtained through structure from motion. The full real plant models in the ROSE-X data set together with the synthetic data we employed in this work can be greatly instrumental for a systematic analysis of the responses of the architectures to low quality and noisy 3D data through simulation of acquisition systems such as ToF cameras and LiDARs in virtual environments Also, data augmentation is possible by introducing variable point density and artificial noise to the point clouds. However, the architectures should eventually be tested on data acquired by low-cost systems including structure from motion.

Another issue is that the variability of local parts is greatly effected by the intricate plant structure, bringing distinct parts close to each other. The training data should be able to account for diverse local geometric occurrences, such as touching leaves or branches due to dense foliage. More realistic synthetic data or plant-specific augmentation techniques ensuring folding of leaves and branches can help enrich the labeled data.

\section{Conclusion}

We modified six recent 3D point-based deep learning architectures, PointNet, PointNet++, DGCNN, PointCNN, ShellNet, and RIConv, for segmentation of 3D models of real rosebush plants into their structural parts. We used the annotated 3D models in ROSE-X data set for training and testing the networks. We also conducted experiments where the networks were pre-trained with synthetic rosebush models generated by L-studio software, and then updated by real rosebush data. The results indicate that pre-training with synthetic data boosts the performance of all networks, except PointNet. The best segmentation results were obtained by PointNet++ with a mean $IoU$ rate of 86.19\%. We attribute this success to the ease of determining the size of the hierarchical local regions to extract multi-scale features with PointNet++. RIConv was not as effective due to reliance on rotation invariant features that provide insufficient local geometric information. DGCNN , PointCNN, and ShellNet produced promising results, however defining local regions for feature extraction by K-neighbourhood of points is less practical for modeling plant geometry; since the optimum K for each scale will be dependent on both the size of the plant part structures and the point density of the 3D point cloud.

\bibliographystyle{elsarticle-num-names}

\bibliography{PointBasedDeepLearning2.bib}

\end{document}